\documentclass[runningheads]{llncs}
\usepackage{graphicx}
\usepackage{orcidlink}
\usepackage{booktabs}
\usepackage{amsmath, amsfonts}
\usepackage{comment}
\usepackage{multicol}
\usepackage{multirow}
\usepackage[super]{nth}
\usepackage{placeins}
\usepackage{enumitem}

\usepackage{caption}
\usepackage{subcaption}

\let\originalleft\left
\let\originalright\right
\renewcommand{\left}{\mathopen{}\mathclose\bgroup\originalleft}
\renewcommand{\right}{\aftergroup\egroup\originalright}

\begin{document}
\title{Beyond the Veil of Similarity: Quantifying Semantic Continuity in Explainable AI}
% Beyond the Veil of Similarity: Quantifying Semantic Continuity in Explainable AI ?
%
\titlerunning{Quantifying Semantic Continuity in XAI}

\author{Qi Huang\inst{1}\orcidlink{0009-0007-4989-135X} \and Emanuele Mezzi\inst{2}\orcidlink{0009-0001-9007-8260} \and Osman Mutlu\inst{3}\orcidlink{0000-0001-6144-5685}  \and Miltiadis Kofinas\inst{4}\orcidlink{0000-0002-3392-4037} \and Vidya Prasad\inst{5}\orcidlink{0000-0002-9296-3693} \and Shadnan Azwad Khan\inst{6}\orcidlink{0000-0003-2769-6856} \and Elena Ranguelova\inst{7}\orcidlink{0000-0002-9834-1756} \and Niki van Stein\inst{1}\orcidlink{0000-0002-0013-7969} 
}
\authorrunning{Q. Huang et al.}
\institute{\textit{Leiden University}, Leiden, The Netherlands \\
\email{\{q.huang,n.van.stein\}@liacs.leidenuniv.nl}
\and
\textit{Vrije Universiteit Amsterdam}, Amsterdam, The Netherlands
\and
\textit{Wageningen Food Safety Research}, Wageningen, The Netherlands
\and
\textit{University of Amsterdam}, Amsterdam, The Netherlands
\and
\textit{Eindhoven University of Technology}, Eindhoven, The Netherlands 
\and
\textit{Sorbonne University}, Paris, France
\and
\textit{Netherlands eScience center}, Amsterdam, The Netherlands
}
\maketitle              % typeset the header of the contribution
\begin{abstract}
We introduce a novel metric for measuring semantic continuity in Explainable AI methods and machine learning models. We posit that for models to be truly interpretable and trustworthy, similar inputs should yield similar explanations, reflecting a consistent semantic understanding. By leveraging XAI techniques, we assess semantic continuity in the task of image recognition. We conduct experiments to observe how incremental changes in input affect the explanations provided by different XAI methods. Through this approach, we aim to evaluate the models' capability to generalize and abstract semantic concepts accurately and to evaluate different XAI methods in correctly capturing the model behaviour. This paper contributes to the broader discourse on AI interpretability by proposing a quantitative measure for semantic continuity for XAI methods, offering insights into the models' and explainers' internal reasoning processes, and promoting more reliable and transparent AI systems.

\keywords{
Semantic Continuity \and
Explainable AI \and
Machine Learning Interpretability \and
Semantic Analysis}
\end{abstract}

\section{Introduction}

Human intelligence can project objects and events to higher-order semantics. Starting from concrete objects it then generates abstractions that are invariant to the change of the environments in which those objects were initially identified. Over the years, given the growing power of representation shown by deep learning (DL) models, specifically deep neural networks (DNNs), researchers have started to wonder whether this capacity to abstract is only exclusive to humans or also embedded in the neural architecture.
% One of the most important fields in AI through which researchers try to find the answer to this question is represented by Explainable AI (XAI), which aims to enlighten what the features are that contribute to a specific prediction realized by a neural network (or other machine learning model).
The field of Explainable AI (XAI) aims to find the answer to this question, highlighting the features contributing to a specific prediction realized by a neural network (or other machine learning model).

Over time, one of the concepts that characterized XAI called \emph{continuity} has emerged and was highlighted in  \cite{co-12}. It can be framed as the capacity of the XAI method, i.e., the explainer, to behave consistently with the model behaviour. More concretely, the explanations of similar model inputs that result in similar model outputs (confidence) should also correspondingly give similar explanations.

In this work, we investigate \emph{semantic continuity}, which we outline as ``similar semantics should have similar explanations". We first define the scope of this work and a general background in XAI in Section \ref{sec:xai}. We then list related literature and the motivation of our proposed solution in Section \ref{sec:related}. We then define the term semantic continuity in Section \ref{sec:sc} and propose a new metric to measure the semantic continuity of a given model and explainer. In Section \ref{sec:exprimental}, we investigate the semantic continuity in the image domain, and we discuss the results in \ref{sec:results}. Lastly, in Section \ref{sec:conclusion} we summarize our findings and provide our final remarks.

% \textbf{Todos:}
% \begin{itemize}
%     \item Experiment Qi. -- face images - https://genforce.github.io/interfacegan/ - binary tasks
%     \item Experiment Different explainers Text. / KernelSHAP, Lime, Rise. @Shadnan 
%     \item Experiment with different explainer hyper-parameters (Shape). @Vidya Lime, Shap and Gradcam -- send hyperparameters to Osman
%     \item Experiment Different explainers Image. / KernelSHAP, Lime, Rise @Osman
%     \item @Elena will ask for Rise hyperparameters
%     \item Pearson and Kendal Tau for statistical interpretation. @Emanuele look into other measures.
%     \item Writing, Niki will take first shot of existing stuff.
% \end{itemize}

% \newpage
\section{Explainable AI}
\label{sec:xai}
The field of XAI, motivated by the imperative to understand the inner workings of AI models, has undergone many advancements in recent years. This has resulted in various explanation methods being developed that may be broadly categorized into attribution-based, model-based, and example-based explanations \cite{lopes2022xai}. Particularly, attribution-based methods provide insights into the importance of the features in a model by assigning importance values or ranks based on the relevance of these features to final predictions. There are global explanation methods such as Global Sensitivity Analysis \cite{9903639}, which attribute importance to features for a given model on a global level, and there exist various methods for single predictions (local methods), such as perturbation-based, gradient-based, surrogate-based, and propagation-based methods \cite{samek2023explainable}. This discussion mainly focuses on perturbation-based and gradient-based methods for single predictions.

\section{Related Work}
\label{sec:related}
Over the past years, several evaluation frameworks and metrics have emerged to assess the performance and compare different XAI methods with each other. One of these measures is continuity.
Continuity is a critical aspect that ensures the stability and generalizability of XAI solutions and is often associated with robustness. In \cite{cugny2022autoxai}, an AutoXAI framework is proposed that automates the selection of XAI solutions where continuity is a property that plays a significant role. Understanding this property becomes even more crucial in scenarios such as Part-Prototype Image classifiers, where continuity directly influences user trust and the model's ability to generalize \cite{nauta2023co}.

To evaluate the continuity property in XAI solutions, several tools and methodologies have been developed, with some focusing on image-based tasks \cite{ijcai2023p747}. Toolkits designed for XAI evaluation in continuity tests on images include: Quantus \cite{hedstrom2023quantus}, Safari \cite{huang2023safari}, XAI-Bench \cite{xai-bench-2021}, and BAM \cite{BAM2019}. Additionally, more generic XAI toolkits like Captum \cite{kokhlikyan2020captum} and OmniXAI \cite{wenzhuo2022-omnixai} are available. However, the continuity metrics utilized in these tools require further verification.

Examining perturbed inputs has revealed complexities surrounding continuity, particularly when the perturbations lead to misinterpretations and inconsistencies in explanation outcomes. Two types of misinterpretations come to light: one where a perturbed input with different highlighted features receives the same prediction label, and another, where a perturbed input with similar highlighted features is assigned a different prediction label \cite{huang2023safari}.

Additionally, the assumption that explanations and model outcomes are directly comparable has been brought into question. Such as in \cite{Hedstrom1468140}, where experimental demonstrations with image classification tasks, employing various perturbation techniques (Gaussian noise variation, spatial rotation, spatial translation, and latent sampling) are tested, utilizing distance measures such as Maximum Mean Discrepancy and Lipschitz Continuity. The findings support the notion that the outcome is affected by the perturbation techniques and the distance measures employed. It is concluded that the continuity test as it stands, can be easily biased towards desired results by employing a particular combination of perturbation technique and distance measure. This prompts the need for a more rigorous assessment of continuity.

The study in \cite{XAIreliability2021} investigates the effect of Adversarial Perturbations (APs), and subtle disruptions in input data on DNNs. It uses GradCAM to generate explainability maps and shows a decrease in correlation coefficients between Layered GradCAM outputs after a DeepFool attack. Correspondingly, the research in \cite{wu2020experimental} is focused on the problem of semantic discontinuity of deep learning models, where small perturbations in the input space tend to cause semantic-level interference to the model output, which is explained by the flaws in choosing the training targets. The need to study continuous perturbations of the input data is also argued in \cite{kamath2023rethinking}, especially for model interpretability. The authors have addressed the causes for the previously observed fragility of many attribution XAI methods and propose enhanced metrics and improving robustness via adversarial training. The focus of those and similar papers is only on studying the DL method itself. While adversarial robustness is very important, in the context of explaining the model's behaviour in semantic terms, the potential influence of an explainer has not been researched.

A system for visual analytics and understanding of CNNs, VATUN, is presented in \cite{vatun:2021}. Again, the focus is on studying the sensitivity to and preventing adversarial attacks and is also limited to GradCAM images. Similarly, Perturber, introduced in \cite{CNNrobust:2021}, is a web application offering interactive comparison of base and adversarially trained CNN models. Our proposed approach is generic for any data modality, DL model architecture, and attribution explainer. Also, both VATUN and Perturber are interactive approaches and do not include a metric for quantifying (semantic) continuity, which we propose in this work.

As far as we know, here we are the first to define the notion of ``semantic continuity" in the XAI context. In the inspirational publication \cite{co-12}, $12$ properties of explainers have been defined for their objective and systematic evaluation. Continuity considers how continuous is the explanation function learned by the explainer. A continuous function ensures that small variations in the input lead to small changes in the explanation. Continuity also adds to generalizability beyond a particular input, which is specifically useful for domain experts who are not (X)AI experts. Having primarily their needs in mind, we have extended the notion of continuity to semantic continuity as a valuable property of an explainer.

\section{Semantic Continuity}\label{sec:sc} 
% Let \(\theta\) be the semantic change in the input, \(C\) the result of the model, and \(E(C)\) the explanation. Given the assumption that the explainer achieves high correctness, the following should apply if the model is semantically continuous:

% \begin{itemize}
%     \item A discrete change in \(\theta\) in the input should have a change in the result of the model and thus the explanation as well.
%     \item Given two consecutive values of \(\theta\) in the input, the change in the explanation should be continuous in regard to the change of explanations for the previous consecutive values. 
%     \item Given the continuous change in \(\theta\) in the input as the x, the resulting change in the explanation should be monotonic.
%     \item The monotonicity of this function can be measured using Kendall's Tau to arrive at a single metric for the semantic continuity of the model.
% \end{itemize}
To define semantic continuity, we first look at the definition of continuity in the context of explainable AI from the work \cite{co-12}.
The definition of continuity in \cite{co-12} is loosely defined as ``similar inputs should have similar explanations". We expand this idea with the notion of ``semantic continuity" so that ``\textit{semantically} similar inputs should have similar explanations." This definition brings us to the following hypothesis: 

\begin{itemize}[leftmargin=*]
    \item A slight change in the input will correspond to a slight change in the output, which is the result of the explainer.
    \item Given the explanation of a reference prediction (the base case), the bigger the input change, the bigger the change between the explanation of the current output and the explanation of the initial (reference) output. This will yield an increasing monotonic correlation between changes in explanations and changes in the original input (images).
\end{itemize}

\begin{definition}
\label{definition: semantic-XAI}
% Let \(\theta\) be the change in the input, \(C\) the result of the prediction from a Deep Learning model and \(E(C)\) the output of the explainer, we can define semantic continuity with the following system of equations,
Let \(\mathbf{x}_0\) denote the reference input data,
and \(f\) be a function that applies a semantic variation \(\theta\) with a domain \(\mathrm{\Theta}\) on the input data,
resulting in \(\mathbf{x}_i = f(\mathbf{x}_0; \theta_i)\).
Let \(M\) be a deep learning model and \(E(M)\) the explainer of the model. We define semantic continuity as follows: 
\begin{align}
    \label{eq:equation}
    % \begin{cases}
        % \theta_{j} - \theta_{i} = \theta_{k} - \theta_{j} \Rightarrow E(C_{j}) - E(C_{i}) = E(C_{k}) - E(C{j}) \\
    %     \theta_{j} - \theta_{1} > \theta_{i} - \theta_{1} \Rightarrow E(C_{j}) - E(C_{1}) > E(C_{i}) - E(C_{1})
    % \end{cases}
    % \theta_{j} - \theta_{0} > \theta_{i} - \theta_{0} \Rightarrow E(C_{j}) - E(C_{0}) > E(C_{i}) - E(C_{0})\ \forall \theta \in \mathrm{\Theta},
    \theta_{j} - \theta_{0} > \theta_{i} - \theta_{0} \Rightarrow \mathrm{D}\left(E(M(\mathbf{x}_{j})), E(M(\mathbf{x}_{0}))\right) > \mathrm{D}\left(E(M(\mathbf{x}_{i})), E(M(\mathbf{x}_{j}))\right),
\end{align}
\end{definition}
% where \(\theta_0\) and \(E(C_{0})\) correspond to a reference semantic variable and its explanation, respectively.
\(\forall \theta_i, \theta_j \in \mathrm{\Theta}\), where \(\theta_0\) corresponds to an identity transformation, \emph{i.e.} \(\mathbf{x}_0 = f(\mathbf{x}_0; \theta_0)\) and \(E(M(\mathbf{x}_{0}))\) corresponds to its explanation. The function \(D\) corresponds to a distance function between the two explanations.

To test whether XAI methods are semantically continuous, given that the predictor is a perfect predictor that is also semantically continuous, and thus respects the mathematical definition on which we ground the concept of semantic continuity, we propose the following approach: First, given a trained predictor (for example a DL model) for a boolean classification task, we define one or more semantic variations we can apply to the data. For example in the case of image classification, we can apply image transformation techniques such as rotation, contrast change and cropping, which do not alter the semantic meaning of the image. On the other hand, we can also define a gradual transformation from one class to the other class (and therefore gradually changing semantics). Next, we apply the predictor to predict images generated using the semantic variations and subsequently apply an explainer to get the feature attribution map for the prediction. Finally, we can measure the distance between the original (non-transformed) image and the semantic variation of the input image and we can measure the distance between the two feature attribution maps given by the explainer. These distances should increase monotonically when applying larger semantic variations and the distances between inputs should be correlated to distances of the explanations derived by the explainer.

% \begin{enumerate}
%     \item Select a simple binary classification use case.
%     \item Train a neural network to perform binary classification.
%     \item Apply XAI methods to derive the features on which the model bases its predictions.
%     \item Derive a mathematical function that describes semantic continuity and check if it respects the property (\ref{eq:equation}). 
% \end{enumerate}
% select a simple binary classification use case, train a neural network to perform a binary classification, apply explainers to derive which are the features on which the model bases its predictions, derive a mathematical function that describes semantic continuity and check if it respects the property (\ref{eq:equation}). 

\subsection{Proof-of-concept Experiment}
To investigate the semantic continuity of XAI methods, the simplest case study consists of binary classification. We will first explore semantic continuity by considering the case in which the machine learning model must distinguish between triangles and circles, where the grayscale images contain only one uniform triangle or circle positioned in the centre of a uniform background. We selected this case study as proof of concept as the task is simple and can be easily manipulated. In addition, we can assume (and verify) that the model on this task is more or less a perfect predictor. 

Considering that the case study is a binary classification task, we build a Convolutional Neural Network (CNN) of $2$ hidden layers, which is sufficient for a model to learn the features that enable it to distinguish between triangles and circles ($100\%$ test accuracy).

\subsubsection{Generation of Datasets with Semantic Variations}
\begin{figure}[]
    \vspace{-1em}
    \centering
    \includegraphics[width=0.98\textwidth]{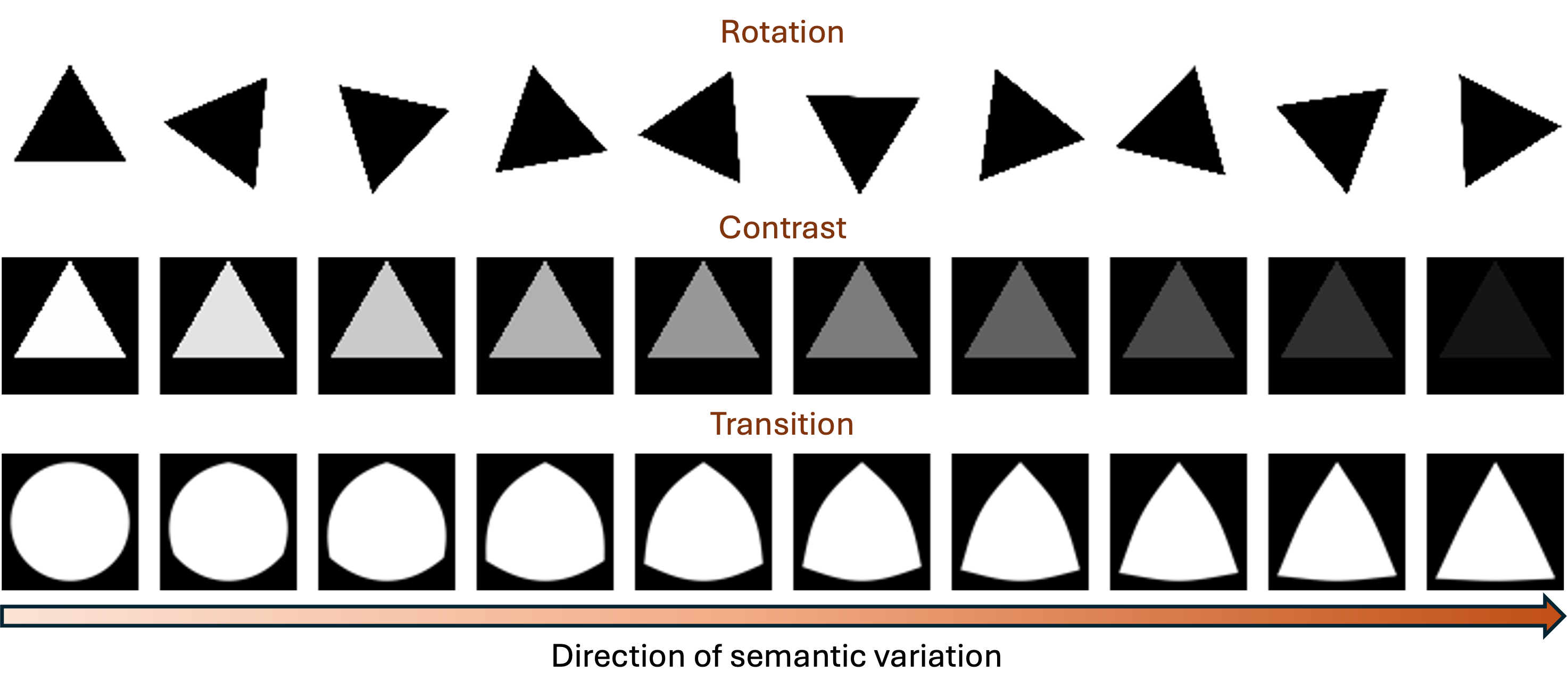}
    \caption{Demonstrations of the data used for the proof-of-concept experiment.}
    \label{fig:shape-examples}
\end{figure}

Once the model has been trained, we generate datasets that enable us to measure the semantic continuity of XAI methods in different scenarios. As shown in Figure~\ref{fig:shape-examples}, we analyze three possible cases of semantic variation and continuity: 

\begin{itemize}
    \item Rotation: the explainer is semantically continuous concerning the rotation of triangles. To check this property, we generate a dataset of $100$ images of the same triangle on the same background. The dataset is a sequence of images with the triangle rotating clockwise by one degree.
    \item Contrast: the explainer is semantically continuous concerning variations in the background contrast. To check this property, we built a dataset composed of $200$ images. The first $100$ are images of triangles, and the second $100$ are circles. In this case, the progressive change consists of a constantly diminishing contrast of the shape with the background until the shape is no longer recognizable.
    \item Transition: the most complex semantic transformation. The rotation and contrast are fixed. The dataset, composed of $100$ images, is a sequence where the starting image depicts a circle and the ending image - a triangle. The shape gradually changes from a circle to a triangle in the in-between images.
\end{itemize}
% \vspace{-1em}
\subsubsection{Comparing Different XAI Methods}

The (pre)trained model is used in each of the three scenarios above to measure the semantic continuity of different explainers.
%Once the three scenarios to test semantic continuity have been designed and the datasets are prepared, it is used the trained model on each of the datasets and applied explainers to each model prediction. 

We test the semantic continuity of RISE \cite{petsiuk2018rise}, LIME \cite{ribeiro2016should}, GradCAM \cite{selvaraju2017grad}, and SHAP \cite{lundberg2017unified} explainers, as these are popular and well-established XAI methods. 
In this proof-of-concept experiment, we assume that the model is a perfect predictor and the output of the model adheres to the semantic continuity definition. 
%In the use of the explainers to model predictions and in the successive phase of metric calculation, we assume that the layer of complexity added by the explainer is negligible and thus it allows us to check for the semantic continuity of the model and not of the explainer. 

Based on this assumption, it is possible to check whether it respects the mathematical assumption in Equation (\ref{eq:equation}) through qualitative analysis and quantitative correlations between the changes in input and explainer output. 
% For each dataset, the independent variable represented on the x-axis will be the difference in semantic variation between each image in the dataset and the first (base) image. This difference always grows for contrast and transition variations, while it will be periodical for the rotation. The dependent variable, instead is the distance between the result gathered from the application of the explainer on the first prediction and the result of the application of the explainer on the prediction of successive images in the dataset. 

To understand the extent to which the XAI output under gradual semantic variations respects monotonicity, the verification is composed of two phases, the first consisting of visual inspection and the second consisting of the application of a correlation metric apt for monotonicity checking to investigate whether a positive change in the input is correlated with a positive change in the explainer output. 
The correlation metrics that we use to quantify semantic continuity are the Pearson \cite{pearson1895notes}, Spearman's \cite{spearman1987proof} and Kendall's Tau \cite{kendall1938new} metric.
%Somers' D \cite{somers1962new}, Distance correlation \cite{szekely2007measuring}, and MultiScale Graph correlation \cite{shen2019distance} metric, 
%while to check for semantic continuity in a more realistic and complex setup using a facial dataset, we limit ourselves to the well-established and well understood Pearson, Spearman's and Kendall's Tau metrics as these showed similar results in the proof-of-concept experiment. 

% \subsection{What outcomes can we expect from the XAI methods evaluated?}
% \begin{itemize}
%     \item A binary heatmap
%     \item A feature importance tensor
%     \item 
% \end{itemize}

% \qi{Please carefully check the following sections 4 and 5.2 that I created, they are generalized from our previous study (both in "theory" and practice) and may need proof-read for logical correctness:)}
\subsection{From \textit{Perfect} Predictor to \textit{Imperfect} Predictor}

As the first proof-of-concept experiment relies on a perfect predictor, we refine our definition of semantic continuity to be able to verify semantic continuity of the XAI method not only when the model is a perfect predictor but also when the model output is not semantic continuous.
In addition, the binary classification of geometric shapes is fairly straightforward for the classifier being evaluated, and we would like to show the applicability of our proposed approach in a more realistic real-world setting. 
It is noteworthy that in Definition~\ref{definition: semantic-XAI}, we aim to directly establish a connection between the semantic changes in input data and the changes in post hoc explanations generated by XAI methods. However, recall one of the fundamental requirements of XAI methods, correctness, which aims to achieve high faithfulness of explanations w.r.t. the to-be-explained model~\cite{co-12}. Our implicit assumption for Definition~\ref{definition: semantic-XAI}, is the classifier can consistently perceive the semantic changes in \textbf{test time} such that it is feasible to evaluate the semantic continuity using XAI methods. In the next evaluation scenario, this assumption or constraint is lifted. Following a similar vein to the previous proof-of-concept Shape datasets, we consider the binary classification of human facial information, which is a much more challenging task for the machine learning model, and hence, the XAI explainers are exposed to a more noisy situation where the model is not always providing accurate predictions. The concept and the criterion of semantic continuity are now re-formulated.

\begin{definition}[Semantic variation]
\label{definition: semantic variation}
Let \(\theta\) denote a semantic variation in the range \([\mathrm{\Theta}_A, \mathrm{\Theta}_B], \mathrm{\Theta}_A < \mathrm{\Theta}_B\).
% Let \(\mathbf{x}_0\) denote an input data from a semantic \(\mathrm{\Theta}_A\), and suppose there exists another non-overlapping semantic \(\mathrm{\Theta}_B\), such that \(\mathrm{\Theta}_A \cap \mathrm{\Theta}_B = \emptyset\).
% Consider a variation \(\theta\) from semantic \(\mathrm{\Theta}_A\) to semantic \(\mathrm{\Theta}_B\).
We let \(f\) be a deterministic function that applies such variations \(\theta\) on \(\mathbf{x}_0\), resulting in \(\mathbf{x}_i = f(\mathbf{x}_0; \theta_i)\), where \(\theta_i\) is a real number that quantifies the scale of such variations.
Here, the function \(f\) and the variation indicator $\theta_i$ firstly satisfies:
\begin{align}
\mathrm{P}(H(\mathbf{x}_i)=\mathrm{\Theta_A}\mid x=\mathbf{x}_i) + \mathrm{P}(H(\mathbf{x}_i)=\mathrm{\Theta_B}\mid x=\mathbf{x}_i) = 1,
\end{align} where \(H(\cdot)\) is a fixed (hypothetically) perfect semantic percipient, and \(\mathrm{P}\) is the probability symbol. Secondly, for any pairs of valid \((\theta_i, \theta_j)\), the following causal property shall be held:
\begin{align}
    \theta_{j}> \theta_{i} \Rightarrow \mathrm{P}(H(\mathbf{x}_j)=\mathrm{\Theta_B}\mid x=\mathbf{x}_j) > \mathrm{P}(H(\mathbf{x}_i)=\mathrm{\Theta_B}\mid x=\mathbf{x}_i).
\end{align}

\end{definition}

Definition~\ref{definition: semantic variation} establishes the concept for a data-dependent, unique, deterministic, and controlled semantic variation process between the two semantics (or domains or concepts). Based on this definition, we now provide formal definitions of semantic continuity for both predictors and XAI methods (explainers). 

Let \(M(x)\) be a machine learning model that determines the semantics for an input \(x\), and a post-hoc explainer \(E(M; x)\) of \(M\). Notably, we only consider explainers that implicitly or explicitly produce a real-valued heatmap over the entire \(x\) or a feature importance score for each element in \(x\).
\begin{definition}[Predictor semantic continuity]
\label{definition: continuity-model}
Given a reference data point \(\mathbf{x}_0\) of semantic \(\mathrm{\Theta}_A\) and a variation function \(f(\cdot; \theta)\) that can transform $\mathbf{x}_0$ from \(\mathrm{\Theta}_A\) to \(\mathrm{\Theta}_B\) as defined in Definition~\ref{definition: semantic variation}. We say the model \(M\) is semantically continuous between \(\mathrm{\Theta}_A\) and \(\mathrm{\Theta}_B\) shortly, \(\overrightarrow{\mathrm{\Theta}_{A, B}}\), on \(\mathbf{x}_0\) if for any valid pair of \((\theta_i, \theta_j)\):
\begin{align}
    \theta_{j}> \theta_{i} \Rightarrow \mathrm{P}(M(x)=\mathrm{\Theta_B}\mid x=\mathbf{x}_j) > \mathrm{P}(M(x)=\mathrm{\Theta_B}\mid x=\mathbf{x}_i),
\end{align}
where \(\mathbf{x}_k = f(\mathbf{x}_0; \theta_k)\), a semantic variation following Definition~\ref{definition: semantic variation}, and \(\mathrm{P}(M(x)=\mathrm{\Theta_B}\mid x=\mathbf{x}_k)\) denotes the probability (confidence) of \(\mathbf{x}_k\) to be an instance of domain \(\mathrm{\Theta}_B\), estimated by model \(M\). 
\end{definition}

\begin{definition}[Explainer semantic continuity]
\label{definition: continuity-XAI}
Similarly, given with a predictive model \(M\), a semantic variation function \(f(x;\theta)\) as defined in Definition~\ref{definition: semantic variation}, and a reference data point \(\mathbf{x}_0\), we say the explainer \(E\) is \(\overrightarrow{\mathrm{\Theta}_{A, B}}\) regarding \(M\) on \(\mathbf{x}_0\) if for any valid pair of \((\theta_s, \theta_t)\):
\begin{align}
    \label{eq:semantic-model}
    \mathrm{P_B}(M;\mathbf{x}_s) > \mathrm{P_B}(M;\mathbf{x}_t)
     \Rightarrow \mathrm{D}\left(E(M;\mathbf{x}_{s}), E(M;\mathbf{x}_{0})\right) > \mathrm{D}(\left(E(M;\mathbf{x}_{t}), E(M;\mathbf{x}_{0})\right),
\end{align}
where \(\mathbf{x}_k = f(\mathbf{x}_0; \theta_k)\), \(\mathrm{P_B}(M;\mathbf{x}_k)\) is short for \(\mathrm{P}(M(x)=\mathrm{\Theta}_B\mid x=\mathbf{x}_k)\), and \(D(\cdot, \cdot)\) is a distance metric that quantifies the discrepancy between outputs of the explainer \(E\).
\end{definition}

Notably, when testing semantic continuity for explainers as defined in Definition~\ref{definition: continuity-XAI}, we don't assume any size relationship between the paired indicators of variations. Speaking in general, when we relate these new definitions to the previous one, Definition~\ref{definition: semantic-XAI} relies on the assumption that the to-be-explained predictor holds Definition~\ref{definition: continuity-model} almost surely, but will contradict the requirement of \textit{correctness} for XAI methods when this assumption is not upheld.

\subsection{Synthesis of the human facial dataset}
\label{sec: create facial dataset}
With the aforementioned formal definitions of semantic continuity, we propose to design and create a binary classification (class A vs class B) dataset \(\mathcal{S}\) that satisfies Definition~\ref{definition: semantic variation} but maximally prohibits a strong baseline model from being \(\overrightarrow{\mathrm{\Theta}_{A, B}}\) on all reference points in \(\mathcal{S}\).

\begin{figure}[]
    \vspace{-1em}
    \centering
    \includegraphics[width=0.98\textwidth]{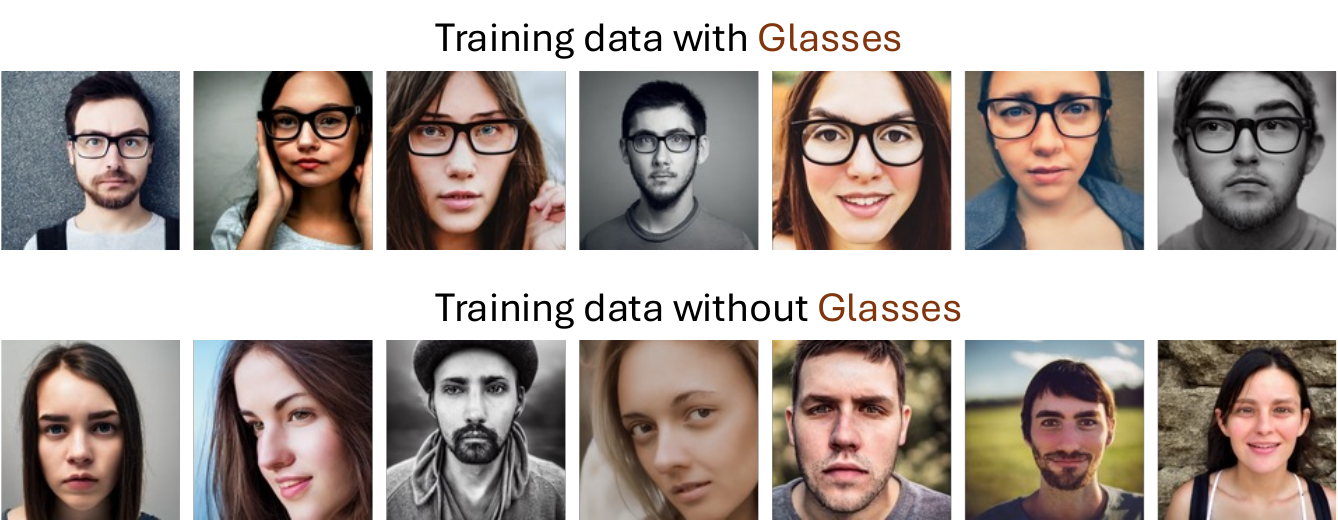}
    \caption{Examples of training data used in our second experiment.}
    \label{fig:face-training-examples}
\end{figure}

\begin{figure}[!ht]
    \vspace{-1em}
    \centering
    \includegraphics[width=0.98\textwidth]{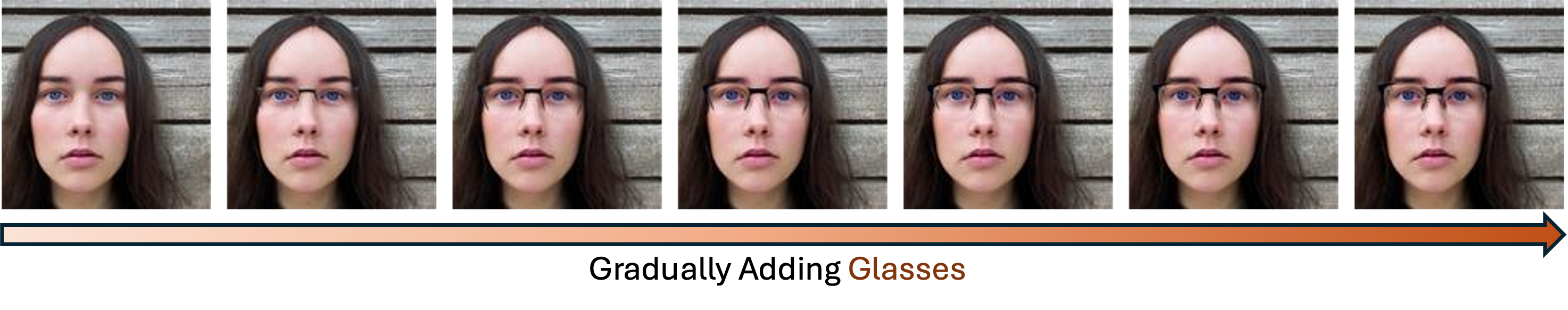}
    \caption{An example sub-series of a test case in our second experiment. The leftmost image illustrates a randomly generated face of a non-real girl without glasses. From left to right, we use generative models to gradually add a pair of \textit{half-rimless} glasses to the images.}
    \label{fig:only-face-example-60916}
\end{figure}

In this research, particularly, we consider classification on artificially generated human faces, where the two non-overlapping classes are \textbf{with glasses} and \textbf{without glasses}. The training data is generated using stable diffusion~\cite{Rombach_2022_CVPR_stable_diffusion}, and all unrealistic samples are manually removed. Figure~\ref{fig:face-training-examples} depicts several randomly chosen examples of our training data.
% \vspace{-1em}
The test dataset is constructed using InterFaceGAN~\cite{shen2020interfacegan} and SEGA~\cite{brack2023sega}. Both are generative models that are capable of smoothly transforming a source image of one semantic to an image of another non-overlapping target semantic while preserving other semantics of the source image during the transformation. The transformation process can be exclusively controlled by a real-valued indicator showing the degree of likeness to the target semantic. Once we determine a sequence of such variation indicators and a starting reference image, it is feasible to generate test data that follows our Definition~\ref{definition: semantic variation}, where the generative model itself serves as both the variation function \(f\) and the semantic percipient \(H\). We give an example of a series of consistently generated test images in Figure~\ref{fig:only-face-example-60916}, where from left to right, the generative model gradually adds a pair of glasses to the face of a girl.
% \vspace{-1em}

\begin{figure}[!ht]
\vspace{-1em}
    \centering
    \includegraphics[width=0.98\textwidth]{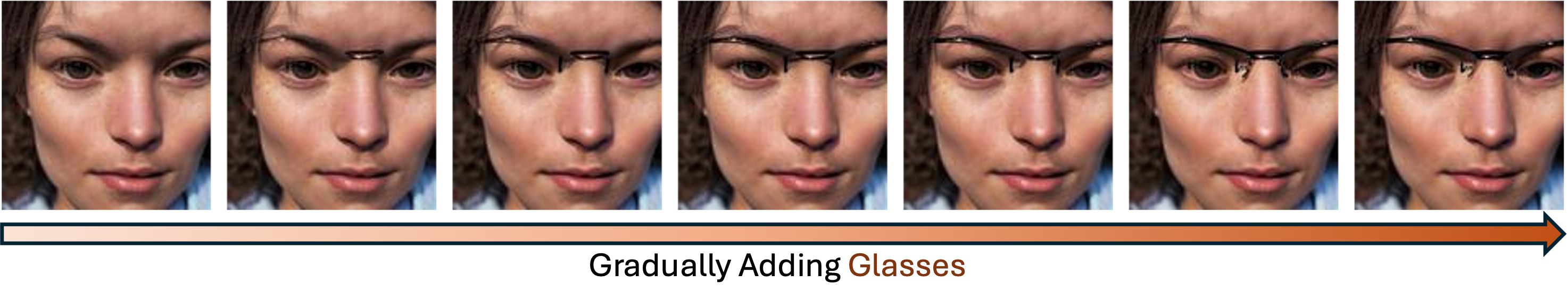}
    \caption{An example sub-series of a test case that does not hold for Definition~\ref{definition: semantic variation}. The context here is the same as that of Figure~\ref{fig:only-face-example-60916}. However, as the semantic variation (indicator) increases, the fidelity of images being \textbf{with glasses} does not increase in the end. And the final image, if talking from human inspections, is not fully convincing to be \textbf{with glasses}.}
    \label{fig:only-face-example-24910}
\end{figure}

Practically, since both generative models are not perfect (regarding humans), it is important to manually inspect their outcomes and discard all series of images that have an unreliable target image as displayed, e.g., in Figure~\ref{fig:only-face-example-24910}. However, we preserve several of these data for analyzing the explainer continuity under the case where Definition~\ref{definition: semantic variation} does not hold for human-level semantic percipient.
\vspace{-1em}
\section{Experimental Setup}
\label{sec:exprimental}

\subsection{Shape Dataset}
\label{sec:shapes}
To explore the (easier to analyse) proof-of-concept scenario related to the capacity of the explainers to capture the semantic continuity in images, we prepare three gray-scale image datasets, each related to a different semantic scenario:
 
\begin{itemize}
    \item Rotation: fixing the background contrast, this dataset contains $100$ images of equilateral triangles with different degrees of rotation. Starting from a base triangle, we apply a progressive rotation to the shape, allowing it to complete a $120$-degree rotation.
    \item Contrast: fixing the rotation, this dataset contains $100$ images of triangles and $100$ images of circles with different background contrast. Starting from the base case of a triangle and a circle with maximum gray-level contrast, we progressively diminish the contrast with the background resulting in a shape that is indistinguishable from the environment.
    \item Transition: fixing rotation and contrast, this dataset contains x images of shapes. Starting from a triangle, the images show a progressive transition until the final image is one of a circle. 
\end{itemize}

The model used \cite{orig-geom-shapes-model} to test the semantic continuity of XAI methods, tested with the above datasets, has been trained on the Simple geometric shapes dataset \cite{orig-geom-shapes-data}. 

\subsection{Synthesis facial dataset}
\label{sec: facial dataset setup}
In section~\ref{sec: create facial dataset}, the methodology for creating the dataset
has been introduced. In total, the training data contains $1000$ balanced samples and the test set contains $100$ balanced samples. Among the test samples, $48$ samples with the label \textbf{no glasses} are chosen to be the reference image for semantic variations, where each of them is gradually shifted towards the \textbf{with glasses} class twenty times uniformly. This results in $1,008$ images for testing the explainers' semantic continuity.

Regarding the classifier, we choose the well-known baseline model ResNet~\cite{he2015deep} with $18$ convolution layers and a Sigmoid output layer. The model is fitted exclusively on the training data from scratch using binary cross entropy loss and Adam optimizer~\cite{kingma2017adam} for about $20$ epochs. As a result, the ResNet-18 achieves $100\%$ predictive accuracy on the $100$ test samples.

For consistency across the setups between experiments, following setups for the Shape dataset, we evaluated instance-wise semantic continuity of RISE, LIME, GradCAM, and KernelSHAP explainers. We choose mean squared deviation and Wasserstein distance as the distance metric (regarding \(D_I)\) for Definition~\ref{definition: continuity-XAI}) in quantitative analysis.

% \subsection{Text Dataset} @Shadnan (@Osman)

% A simplified model and dataset from \cite{orig-stanford-movie-dataset}, featuring only two classes (positive and negative sentiments) has been used as a basis of the Text dataset as it provides a positivity score for all corpus words (adjectives).

\subsection{Software}
For the experiments, the uniform implementation of the XAI explainers in the {Deep Insight and Neural Network Analysis (DIANNA)} \cite{Ranguelova2022}, \cite{Ranguelova_dianna} python library has been used.

\section{Results}
\label{sec:results}

The results section is organized into two subsections: i) results and insights from the proof-of-concept shape dataset to show the outcome of comparisons of different explainers and correlation metrics and ii) results of semantic continuity of explainers on the complex facial image classification task with realistic images. 
%The line plots show the correlation between the semantic change and the explanation change, representing the semantic continuity. So, the sudden changes or spikes in this graph would indicate semantic discontinuity.

% The results for each dataset is shown in a separate figure which contains three scatter plots and a line plot. The scatter plots are the PCA of the inputs put together, PCA of the explanations put together, and the model confidence. The line plot shows the correlation between the semantic change and the explanation change, representing the semantic continuity. So, the sudden changes or spikes in this graph would indicate semantic discontinuity.

\subsection{Proof-of-concept Results: Shape Dataset}

To calculate the extent to which changes in the input lead to changes in the output, we calculate the correlation between the independent variable $x$, which represents the change in the input, and the dependent variable $y$, which represents how explanations vary. 

The distance metrics considered are the Pearson correlation, Spearman's correlation and Kendall's Tau correlation. Considering that the ideal result consists of a monotonic function, that would be able to show the increasing change that affects the output once the input is modified, these correlation metrics have been selected on the basis of their capacity to capture monotonicity. %The comparison of these correlation metrics for each transformation of the shape dataset is shown in Table~\ref{res:metrics_comparison_table}.

For each transformation (contrast, rotation and gradual change from circle to triangle), we perform three types of analysis:
\begin{enumerate}
    \item Saliency maps (heatmaps) of the predictor extracted by the XAI methods. These maps highlight the region of interest in the image for a machine learning model, as inferred by the given explainer.
    \item Relational plots that visualize the correlation among the observed properties:
    \begin{itemize}
        \item \textbf{Semantic variations} measures the degree of semantic variation between the varied images and the reference image.
        \item \textbf{Saliency distances} denotes the mean squared deviation distances between the saliency maps of varied images and that of the reference image.
    \end{itemize}
    \item Statistical correlations between the changes in the input and the changes in distances between the heatmaps. We report Pearson correlation, Spearman's rank correlation, and Kendall rank correlation (Kendall's \(\tau\)) to quantify the degree of explainer continuity.
\end{enumerate}

\begin{figure}[]
\vspace{-1em}
    \centering
    \begin{subfigure}[b]{\textwidth}
    \centering
    \includegraphics[width=\textwidth]{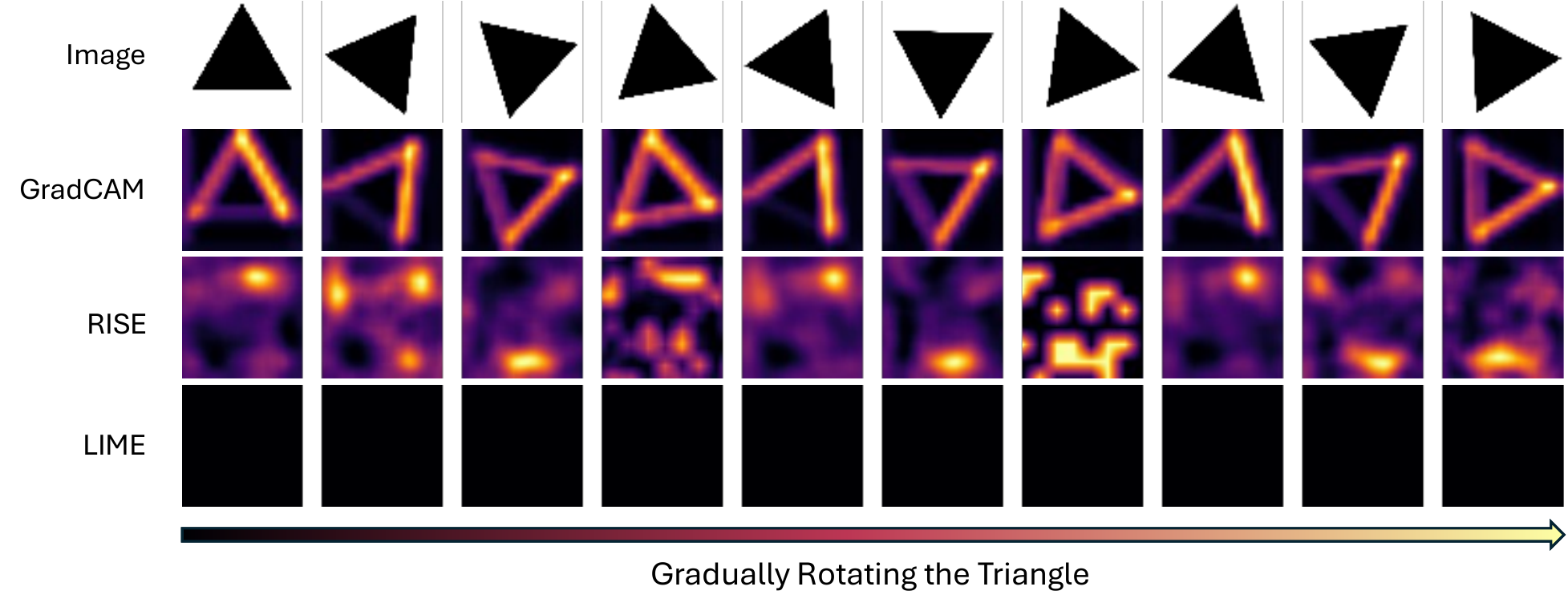}
    \caption{
    The explanations from explainers regarding semantic variations. Darker areas suggest a higher impact on the model prediction.
    % The saliency maps inferred by the four XAI methods regarding semantic variation. Darker areas suggest a higher impact on the model prediction.
    }
    \label{fig:example-rotation}
    \end{subfigure}
    \begin{subfigure}[b]{\textwidth}
    \centering
    \includegraphics[width=\textwidth]{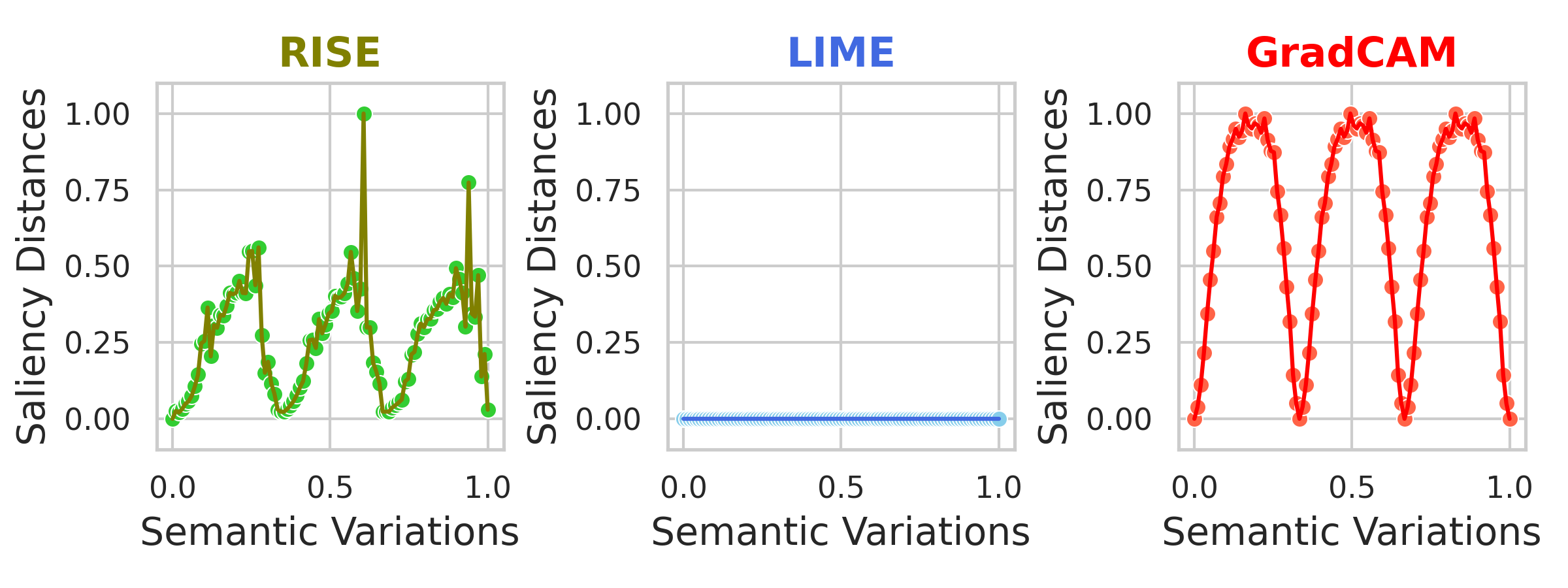}
    \caption{The plots that portray the relation among semantic variation, and saliency distances. All variables are normalized to \([0,1]\) for better visualization. 
    }
    \label{fig:relation-rotation}
    \end{subfigure}
    \begin{subfigure}[b]{\textwidth}
    \centering
    % \begin{tabular*}{\linewidth}{@{\extracolsep{\fill}}ccccc}
    % Correlation           & Metric                           & RISE  & LIME  & GradCAM    \\ \hline
    % \multicolumn{1}{c|}{} & \multicolumn{1}{l|}{Wasserstein} & {\color[HTML]{0000FF} 0.383}     & - & {\color[HTML]{0000FF} -} \\
    % \multicolumn{1}{c|}{\multirow{-2}{*}{Kendall}}  & \multicolumn{1}{l|}{MSD} & 0.833     & - & {\color[HTML]{0000FF} 0.917}                        \\ \hline
    % \multicolumn{1}{c|}{} & \multicolumn{1}{l|}{Wasserstein} & - & - & - \\
    % \multicolumn{1}{c|}{\multirow{-2}{*}{Pearson}}  & \multicolumn{1}{l|}{MSD} & 0.934 & - & {\color[HTML]{0000FF} 0.945} \\ \hline
    % \multicolumn{1}{c|}{} & \multicolumn{1}{l|}{Wasserstein} & -  & - & {\color[HTML]{0000FF} -} \\
    % \multicolumn{1}{c|}{\multirow{-2}{*}{Spearman}} & \multicolumn{1}{l|}{MSD} & 0.932     & - & {\color[HTML]{0000FF} 0.980}                      
    % \end{tabular*}%
    \begin{tabular*}{\linewidth}{@{\extracolsep{\fill}}ccccc}
    Correlation           & Metric                           & RISE  & LIME  & GradCAM    \\ \hline
    \multicolumn{1}{c|}{} & \multicolumn{1}{l|}{Wasserstein} & 0.383     & - & {\color[HTML]{0000FF} 0.733} \\
    \multicolumn{1}{c|}{\multirow{-2}{*}{Kendall}}  & \multicolumn{1}{l|}{MSD} & 0.833     & - & {\color[HTML]{0000FF} 0.967}                        \\ \hline
    \multicolumn{1}{c|}{} & \multicolumn{1}{l|}{Wasserstein} & - & - & {\color[HTML]{0000FF} 0.888} \\
    \multicolumn{1}{c|}{\multirow{-2}{*}{Pearson}}  & \multicolumn{1}{l|}{MSD} & 0.935 & - & {\color[HTML]{0000FF} 0.958} \\ \hline
    \multicolumn{1}{c|}{} & \multicolumn{1}{l|}{Wasserstein} & -  & - & {\color[HTML]{0000FF} 0.865} \\
    \multicolumn{1}{c|}{\multirow{-2}{*}{Spearman}} & \multicolumn{1}{l|}{MSD} & 0.932     & - & {\color[HTML]{0000FF} 0.991}                      
    \end{tabular*}%
    \caption{Statistical correlations between the saliency distances and changes in the input for the first 30 degrees of change. The MSD stands for mean squared deviation, while the Wasserstein is the first Wasserstein distance. The highest correlation coefficient per metric per distance is colored in \textcolor{blue}{blue}. The dash symbol, \textbf{-}, indicates a \(p\geq 0.05\) where it is not evident enough to reject \(H_\mathrm{0}\), i.e., the saliency distances and input changes are likely uncorrelated.\label{tab:stats-rotation}}
    \end{subfigure}
    \caption{
    \textbf{Rotation} transformation.
    }
    \label{fig:result-rotation}
\end{figure}

The analysis of the first transformation (Rotation) can be seen in Figure~\ref{fig:result-rotation}. We can observe that GradCAM focuses on the edges of the triangle and shows a perfect explanation for the triangle class. When looking at the Saliency Distances for GradCAM in Figure~\ref{fig:relation-rotation}, GradCAM shows an oscillating pattern that matches the fact that after $60$ degrees of rotation, the original image is obtained. Also, RISE shows the expected pattern for the rotation case with the exception of a few outliers, although its explanations are less clear. The statistical correlations given in Table~\ref{tab:stats-rotation} match with these observations, note that for these statistics we only look at the first $30$ degrees of change, as for those variations the saliency distances should be monotonically increasing. Lime fails to provide any differences in explanations and can therefore not be calculated. Note that with different hyper-parameters LIME could possibly be improved but that is outside the scope of this work.

\begin{figure}[]
\vspace{-1em}
    \centering
    \begin{subfigure}[b]{\textwidth}
    \centering
    \includegraphics[width=\textwidth]{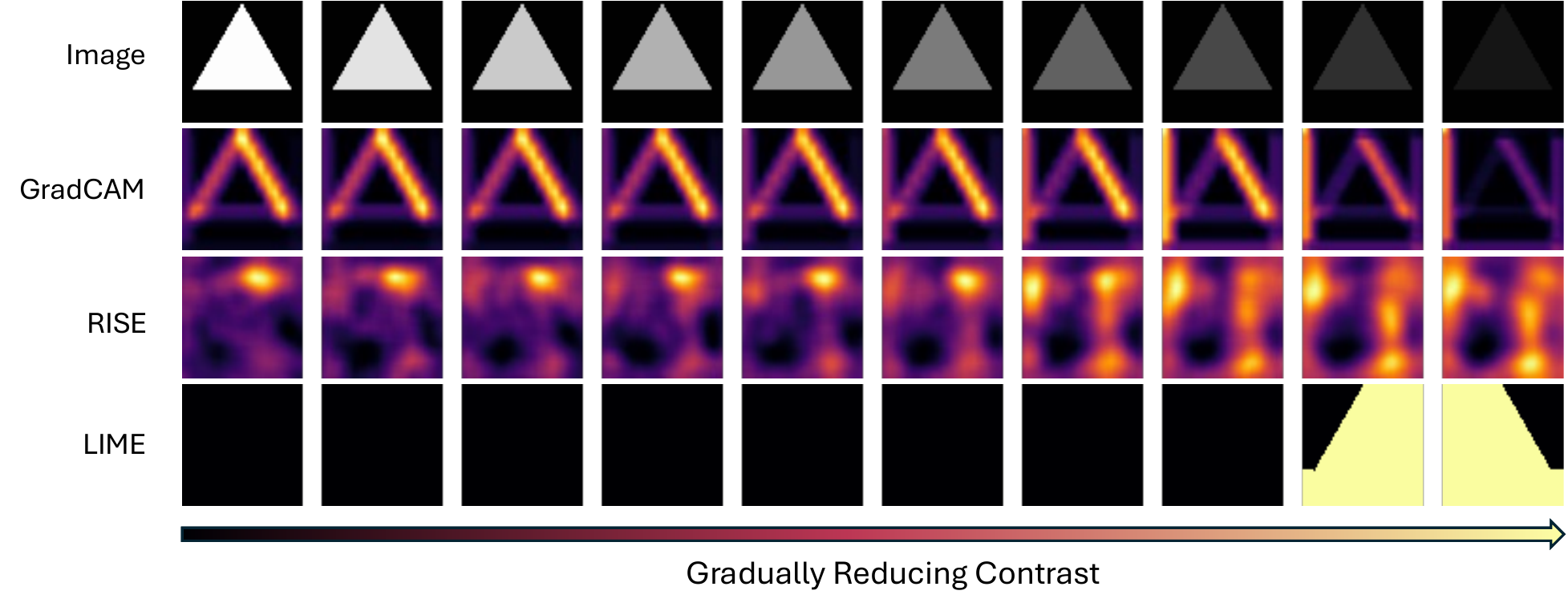}
    \caption{
    The explanations from explainers regarding semantic variations. Darker areas suggest a higher impact on the model prediction.
    % The saliency maps inferred by the four XAI methods regarding semantic variation. Darker areas suggest a higher impact on the model prediction.
    }
    \label{fig:example-color}
    \end{subfigure}
    \begin{subfigure}[b]{\textwidth}
    \centering
    \includegraphics[width=\textwidth]{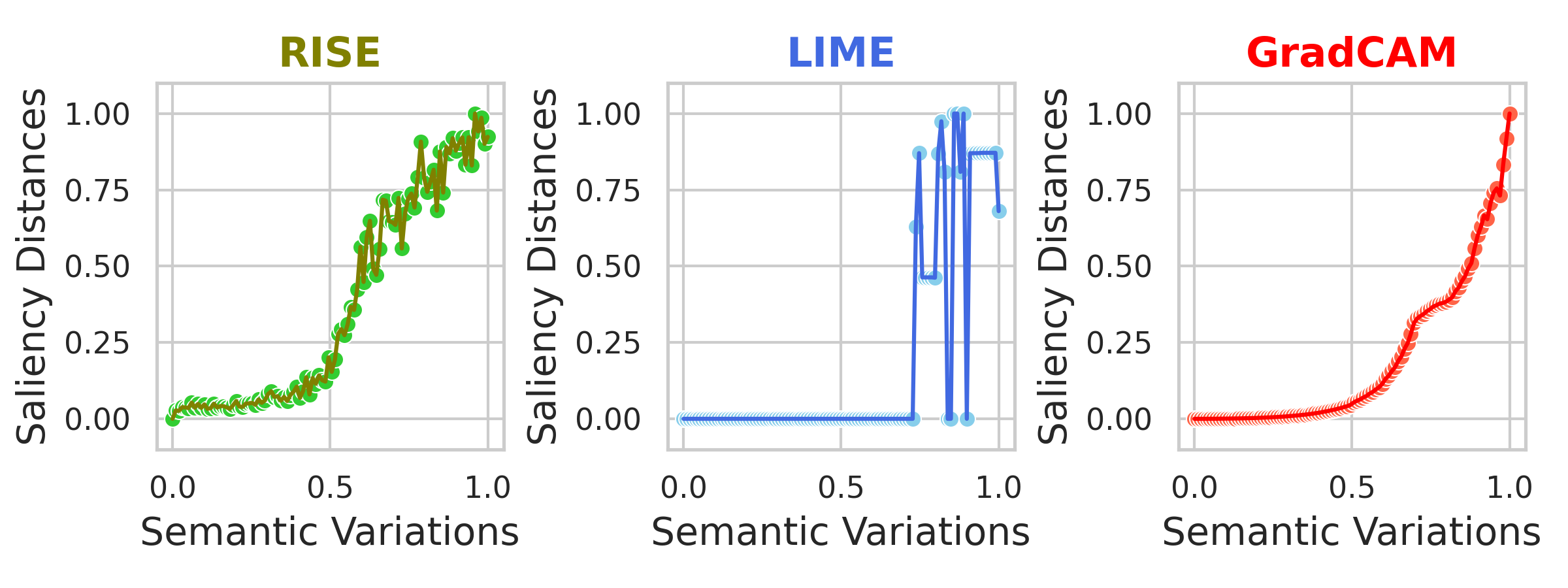}
    \caption{The plots that portray the relation among semantic variation, and saliency distances. All variables are normalized to \([0,1]\) for better visualization. 
    }
    \label{fig:relation-color}
    \end{subfigure}
    \begin{subfigure}[b]{\textwidth}
    \centering
    % \begin{tabular*}{\linewidth}{@{\extracolsep{\fill}}ccccc}
    % Correlation           & Metric                           & RISE  & LIME  & GradCAM    \\ \hline
    % \multicolumn{1}{c|}{} & \multicolumn{1}{l|}{Wasserstein} & {\color[HTML]{0000FF}0.788}     & 0.580 & 0.614 \\
    % \multicolumn{1}{c|}{\multirow{-2}{*}{Kendall}}  & \multicolumn{1}{l|}{MSD} & {\color[HTML]{0000FF}0.887}     & 0.580 &  0.344                        \\ \hline
    % \multicolumn{1}{c|}{} & \multicolumn{1}{l|}{Wasserstein} & {\color[HTML]{0000FF}0.938} & 0.709 & 0.776 \\
    % \multicolumn{1}{c|}{\multirow{-2}{*}{Pearson}}  & \multicolumn{1}{l|}{MSD} & {\color[HTML]{0000FF}0.941} & 0.709 & 0.703 \\ \hline
    % \multicolumn{1}{c|}{} & \multicolumn{1}{l|}{Wasserstein} & {\color[HTML]{0000FF}0.939}  & 0.717 & 0.747 \\
    % \multicolumn{1}{c|}{\multirow{-2}{*}{Spearman}} & \multicolumn{1}{l|}{MSD} & {\color[HTML]{0000FF}0.981}     & 0.717 & 0.553                    
    % \end{tabular*}% 
    \begin{tabular*}{\linewidth}{@{\extracolsep{\fill}}ccccc}
    Correlation           & Metric                           & RISE  & LIME  & GradCAM    \\ \hline
    \multicolumn{1}{c|}{} & \multicolumn{1}{l|}{Wasserstein} & 0.788     & 0.580 & {\color[HTML]{0000FF}0.926} \\
    \multicolumn{1}{c|}{\multirow{-2}{*}{Kendall}}  & \multicolumn{1}{l|}{MSD} & 0.887     & 0.580 &  {\color[HTML]{0000FF}0.999}                        \\ \hline
    \multicolumn{1}{c|}{} & \multicolumn{1}{l|}{Wasserstein} & {\color[HTML]{0000FF}0.938} & 0.709 & 0.891\\
    \multicolumn{1}{c|}{\multirow{-2}{*}{Pearson}}  & \multicolumn{1}{l|}{MSD} & {\color[HTML]{0000FF}0.941} & 0.709 & 0.879 \\ \hline
    \multicolumn{1}{c|}{} & \multicolumn{1}{l|}{Wasserstein} & 0.939  & 0.717 & {\color[HTML]{0000FF}0.979} \\
    \multicolumn{1}{c|}{\multirow{-2}{*}{Spearman}} & \multicolumn{1}{l|}{MSD} & 0.981     & 0.717 & {\color[HTML]{0000FF}1.000}                    
    \end{tabular*}% 
    \caption{Statistical correlations between the saliency distances and changes in the input. The MSD stands for mean squared deviation, while the Wasserstein is the first Wasserstein distance. The highest correlation coefficient per metric per distance is colored in \textcolor{blue}{blue}. The dash symbol, \textbf{-}, indicates a \(p\geq 0.05\) where it is not evident enough to reject \(H_\mathrm{0}\), i.e., the saliency distances and input changes are likely uncorrelated.\label{tab:stats-color}}
    \end{subfigure}
    \caption{
    \textbf{Contrast} transformation.
    }
    \label{fig:result-color}
\end{figure}

The analysis of the contrast transformation can be viewed in Figure~\ref{fig:result-color}. Visually, the explanations of GradCAM seem again to be superior, LIME in this case suffers from its binary nature in providing the explanations. Also, the relation between saliency distances and semantic variations in Figure~\ref{fig:relation-color} and the corresponding correlation metrics in Figure~\ref{tab:stats-color} confirms these observations. In the relation between distances and variations, GradCAM and RISE show the most monotonic pattern (GradCAM seems to be perfectly monotonically increasing), with Kendall's Tau and Spearman giving highest correlation to GradCAM and Pearson correlation is the highest for RISE.

\begin{figure}[]
\vspace{-1em}
    \centering
    \begin{subfigure}[b]{\textwidth}
    \centering
    \includegraphics[width=\textwidth]{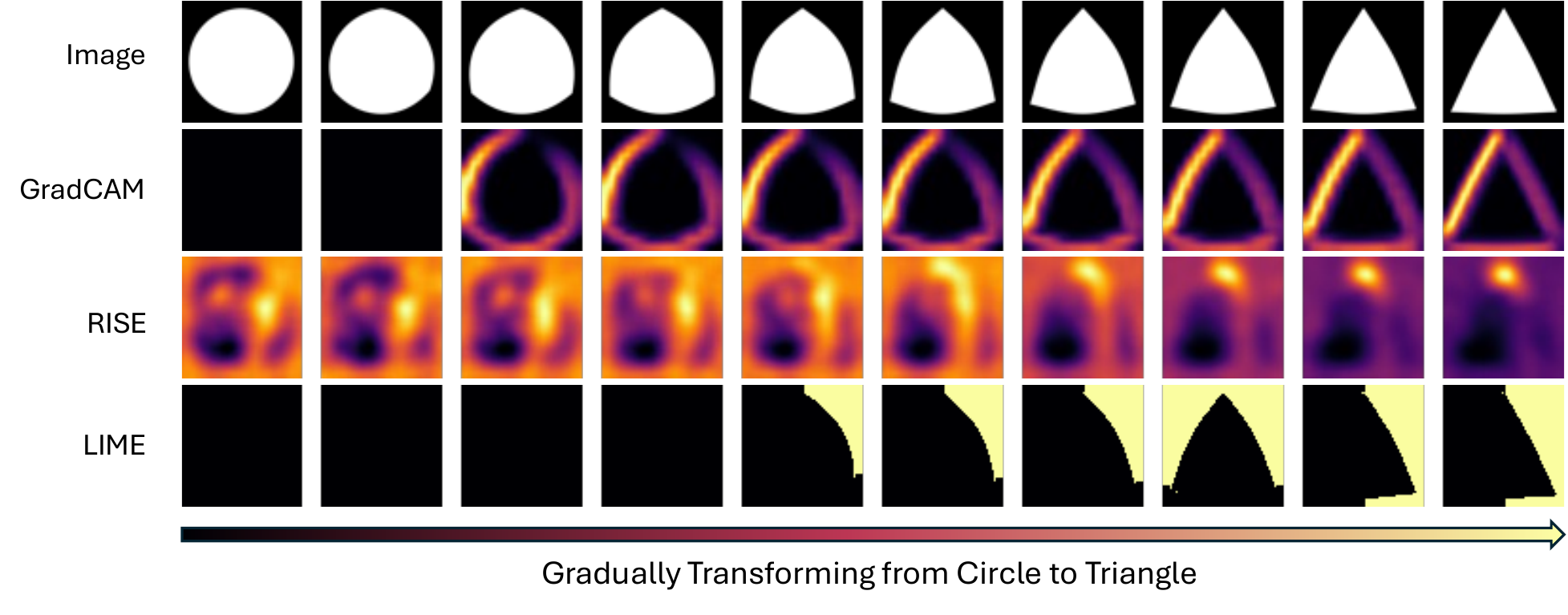}
    \caption{
    The explanations from explainers regarding semantic variations. Darker areas suggest a higher impact on the model prediction.
    % The saliency maps inferred by the four XAI methods regarding semantic variation. Darker areas suggest a higher impact on the model prediction.
    }
    \label{fig:example-roundedness}
    \end{subfigure}
    \begin{subfigure}[b]{\textwidth}
    \centering
    \includegraphics[width=\textwidth]{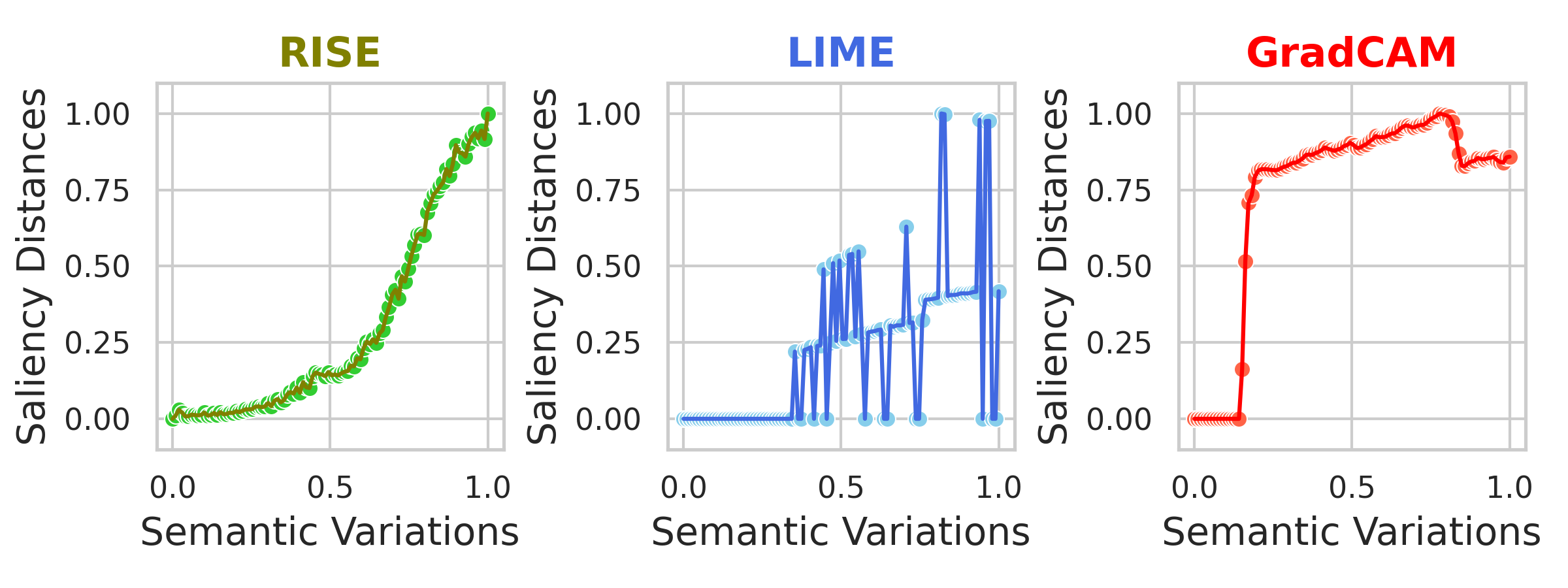}
    \caption{The plots that portray the relation among semantic variation, and saliency distances. All variables are normalized to \([0,1]\) for better visualization. 
    }
    \label{fig:relation-roundedness}
    \end{subfigure}
    \begin{subfigure}[b]{\textwidth}
    \centering
    \begin{tabular*}{\linewidth}{@{\extracolsep{\fill}}ccccc}
    Correlation           & Metric                           & RISE  & LIME  & GradCAM    \\ \hline
    \multicolumn{1}{c|}{} & \multicolumn{1}{l|}{Wasserstein} & {\color[HTML]{0000FF}0.784}     & 0.633 & 0.523 \\
    \multicolumn{1}{c|}{\multirow{-2}{*}{Kendall}}  & \multicolumn{1}{l|}{MSD} & {\color[HTML]{0000FF}0.947}     & 0.633 &  0.586                        \\ \hline
    \multicolumn{1}{c|}{} & \multicolumn{1}{l|}{Wasserstein} & 0.527 & {\color[HTML]{0000FF}0.667} & 0.644 \\
    \multicolumn{1}{c|}{\multirow{-2}{*}{Pearson}}  & \multicolumn{1}{l|}{MSD} & {\color[HTML]{0000FF}0.916} & 0.667 & 0.671 \\ \hline
    \multicolumn{1}{c|}{} & \multicolumn{1}{l|}{Wasserstein} & {\color[HTML]{0000FF}0.909}  & 0.719 & 0.541 \\
    \multicolumn{1}{c|}{\multirow{-2}{*}{Spearman}} & \multicolumn{1}{l|}{MSD} & {\color[HTML]{0000FF}0.947}     & 0.719 &  0.646                      
    \end{tabular*}% 
    % \begin{tabular*}{\linewidth}{@{\extracolsep{\fill}}ccccc}
    % Correlation           & Metric                           & RISE  & LIME  & GradCAM    \\ \hline
    % \multicolumn{1}{c|}{} & \multicolumn{1}{l|}{Wasserstein} & {\color[HTML]{0000FF}0.784}     & 0.633 & 0.708 \\
    % \multicolumn{1}{c|}{\multirow{-2}{*}{Kendall}}  & \multicolumn{1}{l|}{MSD} & {\color[HTML]{0000FF}0.947}     & 0.633 &  0.732                        \\ \hline
    % \multicolumn{1}{c|}{} & \multicolumn{1}{l|}{Wasserstein} & 0.527 & {\color[HTML]{0000FF}0.667} & 0.322 \\
    % \multicolumn{1}{c|}{\multirow{-2}{*}{Pearson}}  & \multicolumn{1}{l|}{MSD} & {\color[HTML]{0000FF}0.916} & 0.667 & 0.269 \\ \hline
    % \multicolumn{1}{c|}{} & \multicolumn{1}{l|}{Wasserstein} & {\color[HTML]{0000FF}0.909}  & 0.719 & 0.699 \\
    % \multicolumn{1}{c|}{\multirow{-2}{*}{Spearman}} & \multicolumn{1}{l|}{MSD} & {\color[HTML]{0000FF}0.947}     & 0.719 &  0.708                      
    % \end{tabular*}% 
    \caption{Statistical correlations between the saliency distances and changes in the input. The MSD stands for mean squared deviation, while the Wasserstein is the first Wasserstein distance. The highest correlation coefficient per metric per distance is colored in \textcolor{blue}{blue}. The dash symbol, \textbf{-}, indicates a \(p\geq 0.05\) where it is not evident enough to reject \(H_\mathrm{0}\), i.e., the saliency distances and input changes are likely uncorrelated.\label{tab:stats-roundedness}}
    \end{subfigure}
    \caption{
    \textbf{Circle to Triangle} transformation.
    }
    \label{fig:result-roundedness}
\end{figure}

The analysis of the circle-to-triangle transformation can be viewed in Figure~\ref{fig:result-roundedness}. Again GradCAM seems to be superior in terms of visualized explanations, however also RISE and LIME show meaningful and logical explanations in this case. In terms of the relation between the saliency distance and semantic variation, GradCAM suffers from the first few explanations being empty and a slight drop in saliency distance towards the end of the transformation, resulting in overall lower correlation measures than RISE for this case.
% For our first comparison, we show the correlation graph of two models with the same structure but using a different random initialization, which were both trained on the shape dataset, side by side in Figure~\ref{res:model_comparison}. The results shown here are from RISE on the contrast task. The results show that different models, even with the same architecture and trained on the same data set, can vary the measured semantic continuity, as the model confidence and behaviour also affect the output of the explainers.

% In Figure~\ref{res:explainer_comparison}, we show the correlation graph of three different explainers, keeping the model and the task persistent. We observe here that both GradCAM and RISE show good semantic continuous behaviour, while LIME tends to be noisy and inconsistent.

% Lastly, in Figure~\ref{res:function_comparison}, we compare two distance functions, Wasserstein distance and Mean Squared Error(MSE) loss, in combination of two different ways to transform a feature attribution heatmap to a value between 0 and 1, minmax and softmax functions. Here we can observe that softmax transformation seems to result in a less noisy correlation between changes to the input and changes to the explainer outputs.

\subsection{Synthesis facial dataset}
In this section, we discuss the semantic continuity of the four evaluated XAI methods, case by case regarding different relationships between the semantic and the ResNet predictor. For each case, we present three types of analysis:
\begin{enumerate}
    \item Saliency maps (heatmaps) of the predictor extracted by the XAI methods. These maps highlight the region of interest in the image for a machine learning model, as inferred by the given explainer.
    \item Relational plots that visualize the correlation among the observed properties:
    \begin{itemize}
        \item \textbf{Semantic variations} measures the degree of semantic variation between the varied images and the reference image.
        \item \textbf{Model confidence} represents the probability of the given image to be of the class \textit{with glasses}. 
        \item \textbf{Saliency distances} denotes the first Wasserstein distances between the saliency maps of varied images and that of the reference image.
        \item \textbf{Confidence changes} quantifies the differences in model confidence between the varied to-be-measured images and the reference image.
    \end{itemize}
    \item Statistical correlations between the changes in model confidence and the changes in distances between the heatmaps. We report Pearson correlation and Kendall rank correlation (Kendall's \(\tau\)) to quantify the degree of explainer continuity as introduced in~\ref{definition: continuity-XAI}. Apart from the first Wasserstein distance, we additionally report the correlation coefficients on saliency distances quantified by mean squared deviation between explanations.
\end{enumerate}
% \qi{I am editing this part to make the tabular statistical results part of figures 9 to 12.}
We discuss our results regarding the four disjoint model's predictive behaviour, commonly referred to as the confusion matrix. For convenience, interchangeably, we use \textbf{G} to denote GradCAM, \textbf{R} for RISE, \textbf{L} for LIME, and \textbf{K} for KernelSHAP. 

\begin{figure}[tp]
\vspace{-1em}
    \centering
    \begin{subfigure}[b]{\textwidth}
    \centering
    \includegraphics[width=\textwidth]{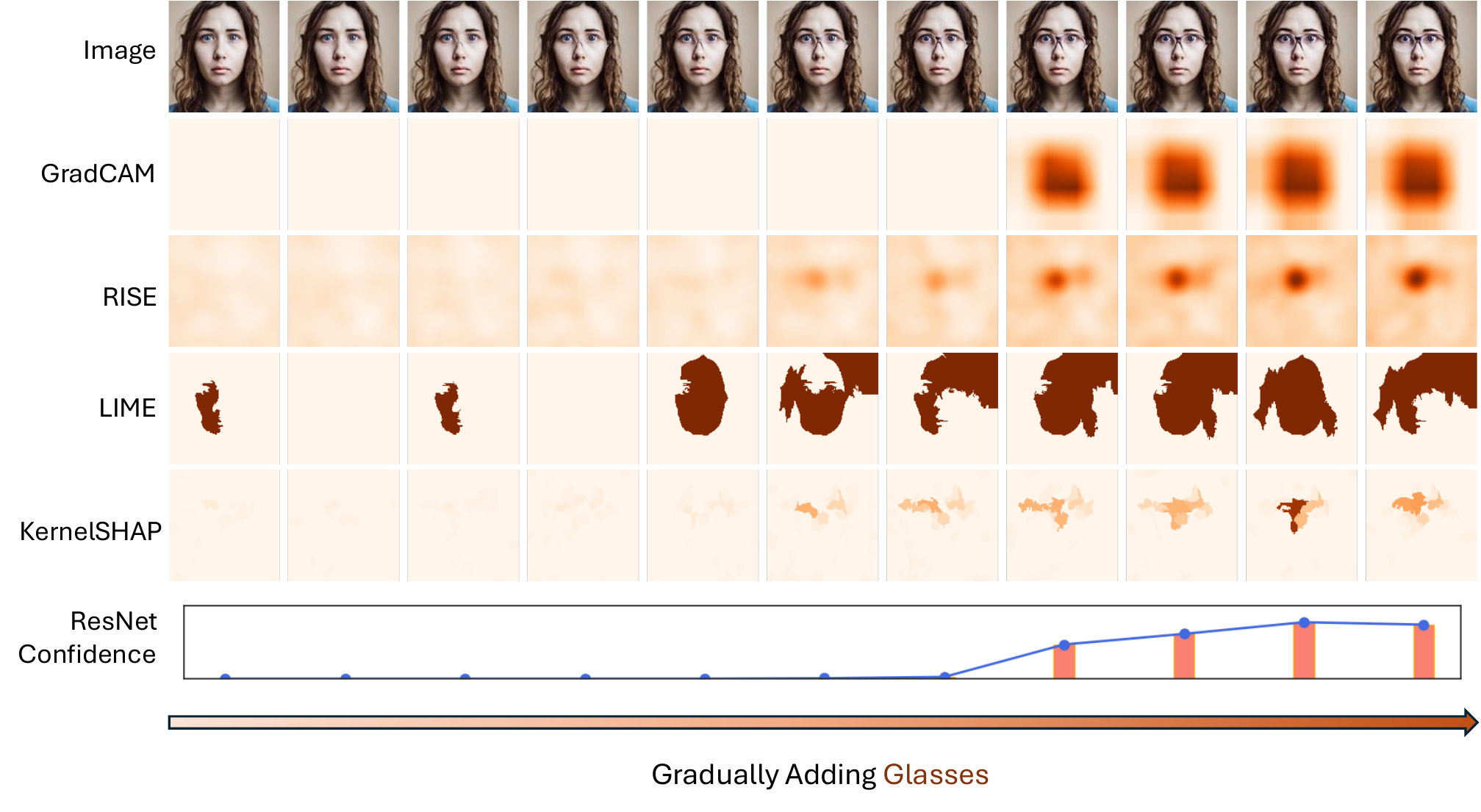}
    \caption{The explanations from explainers and the model confidence regarding semantic variations. Darker areas suggest a higher impact on the model prediction.}
    \label{fig:face-example-9040}
    \end{subfigure}
    \begin{subfigure}[b]{\textwidth}
    \centering
    \includegraphics[width=\textwidth]{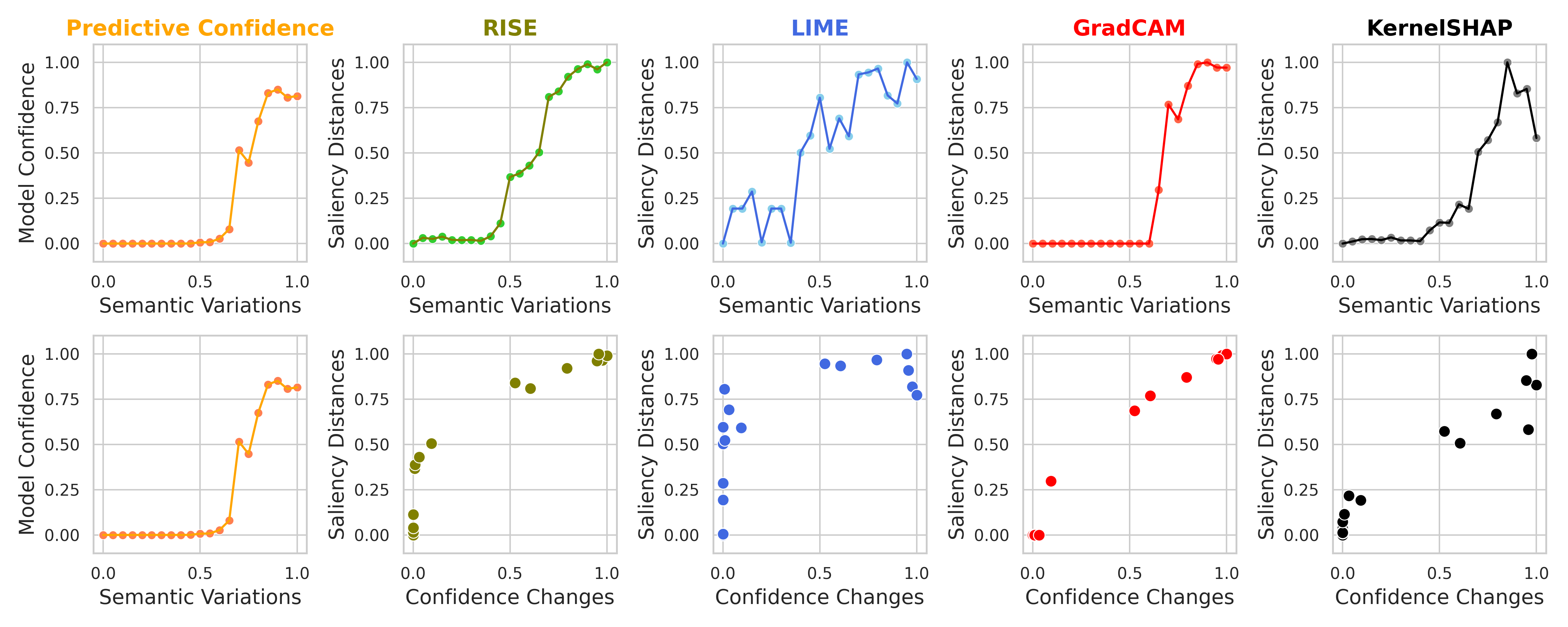}
    \caption{The relation among semantic variation, model confidence, and saliency distances is visualized here. All variables are normalized to \([0,1]\) for better visualization. }
    \label{fig:face-xai-9040}
    \end{subfigure}
    \begin{subfigure}[b]{\textwidth}
        \centering
        \begin{tabular*}{\linewidth}{@{\extracolsep{\fill}}cccccc}
        Correlation Metric                                & Distance                           & RISE  & LIME  & GradCAM & KernelSHAP \\ \hline
        \multicolumn{1}{c|}{\multirow{2}{*}{Kendall}} & \multicolumn{1}{l|}{Wasserstein} & 0.768 & 0.599 & \textcolor{blue}{0.795}   & 0.716      \\
        \multicolumn{1}{c|}{}                         & \multicolumn{1}{l|}{MSD}         & \textcolor{blue}{0.863} & 0.599 & 0.795   & 0.758      \\ \hline
        \multicolumn{1}{c|}{\multirow{2}{*}{Pearson}} & \multicolumn{1}{l|}{Wasserstein} & 0.930 & 0.745 & \textcolor{blue}{0.989}   & 0.716      \\
        \multicolumn{1}{c|}{}                         & \multicolumn{1}{l|}{MSD}         & 0.983 & 0.750 & \textcolor{blue}{0.994}   & 0.758      \\ \hline
        \multicolumn{1}{c|}{\multirow{2}{*}{Spearman}} & \multicolumn{1}{l|}{Wasserstein} & \textcolor{blue}{0.913} & 0.829 & 0.884 & 0.893 \\
        \multicolumn{1}{c|}{}                         & \multicolumn{1}{l|}{MSD}         & \textcolor{blue}{0.965} & 0.820 & 0.884   & 0.908     
        \end{tabular*}%
        \caption{Statistical correlations between the saliency distances and changes in model confidence. The MSD stands for mean squared deviation, while the Wasserstein is the first Wasserstein distance. The highest correlation coefficient per metric per distance is colored in \textcolor{blue}{blue}.
        We choose a threshold \(p<0.05\) to reject \(H_\mathrm{0}\): the saliency distances and changes in model confidence are uncorrelated, and report the statistics.\label{tab:stats-9040}}
    \end{subfigure}
        
    \caption{\textbf{True positive} example of no-glasses to glasses.
    % The explanations from four explainers and the trend of model confidence are visualized in (a), and the relation among semantic variation, model confidence, and saliency distances are given in (b).
    \label{fig:result-9040}}
    %\caption{The predictor ResNet performs similarly to humans in this example, where it correctly predicts the appearance of glasses (a \textbf{true positive}). Speaking intuitively based on the saliency maps, the GradCAM, RISE, and KernelSHAP correctly reflect the changes in model confidence regarding semantic variations as their highlight areas become darker and more concentrated on the locations of glasses. The qualitative relational plots support these observations as GradCAM, RISE, and KernelSHAP possess stably coherent trends with the model's predictive confidence, whereas the plots of LIME show strong stochasticity and do not display strong monotonicity regarding the confidence changes. \label{fig:result-9040}}
    
\end{figure}

\paragraph{True positive}
The first case is a true positive as shown in Figure~\ref{fig:result-9040}. The ResNet model correctly identifies the emergence of glasses of images as they are semantically varied from \textit{no glasses} to \textit{with glasses}. In Figure~\ref{fig:face-example-9040}, we display the saliency maps or the heatmaps produced by four explainers. Speaking intuitively based on the saliency maps, the G, R, and K correctly reflect the changes in model confidence regarding semantic variations as their (particularly R and K) highlight areas become darker and more concentrated on the locations of glasses. The qualitative relational plots in Figure~\ref{fig:face-xai-9040} support these observations as G, R, and K possess stably coherent trends with the model's predictive confidence, whereas the plots of LIME have strong stochasticity and do not display relatively monotonicity regarding the confidence changes. Figure~\ref{tab:stats-9040} depicts the statistical correlations between the saliency distances and the changes in model confidence. With a significance level of 0.05, i.e., \(p<0.05\), we can find both linear and monotonical correlations between the saliency distances of all explainers and the changes in predictor's confidence where GradCAM and RISE are believed to be more semantically continuous regarding the sizes of coefficients.

% \begin{figure}[]
% \vspace{-1em}
%     \centering
%     \includegraphics[width=\textwidth]{figures/face_dataset/Example/pdfs/9040_stack_full.pdf}
%     \caption{The predictor ResNet performs similarly to humans in this example, where it correctly predicts the appearance of glasses (a \textbf{true positive}). The saliency maps inferred by the four XAI methods are also provided accordingly.}
%     \label{fig:face-example-9040}
% \end{figure}
% \vspace{-1em}
% \begin{figure}[]
% \vspace{-1em}
%     \centering
%     \includegraphics[width=\textwidth]{figures/face_dataset/Example/relational_plots/9040_original_changes_Wasserstein_1D_Normalize.png}
%     \caption{The relational plots of Figure~\ref{fig:face-example-9040}. The first column portrays the trend of model confidence regarding the degree of semantic change in inputs. The vertical axes of figures in the second to the last column measure the Wasserstein distances between the saliency maps of varied images and that of the reference image.}
%     \label{fig:face-xai-9040}
% \end{figure}
% \vspace{-1em}
% \vspace{-1em}

\begin{figure}[]
    \centering
    \begin{subfigure}[b]{\textwidth}
    \centering
    \includegraphics[width=\textwidth]{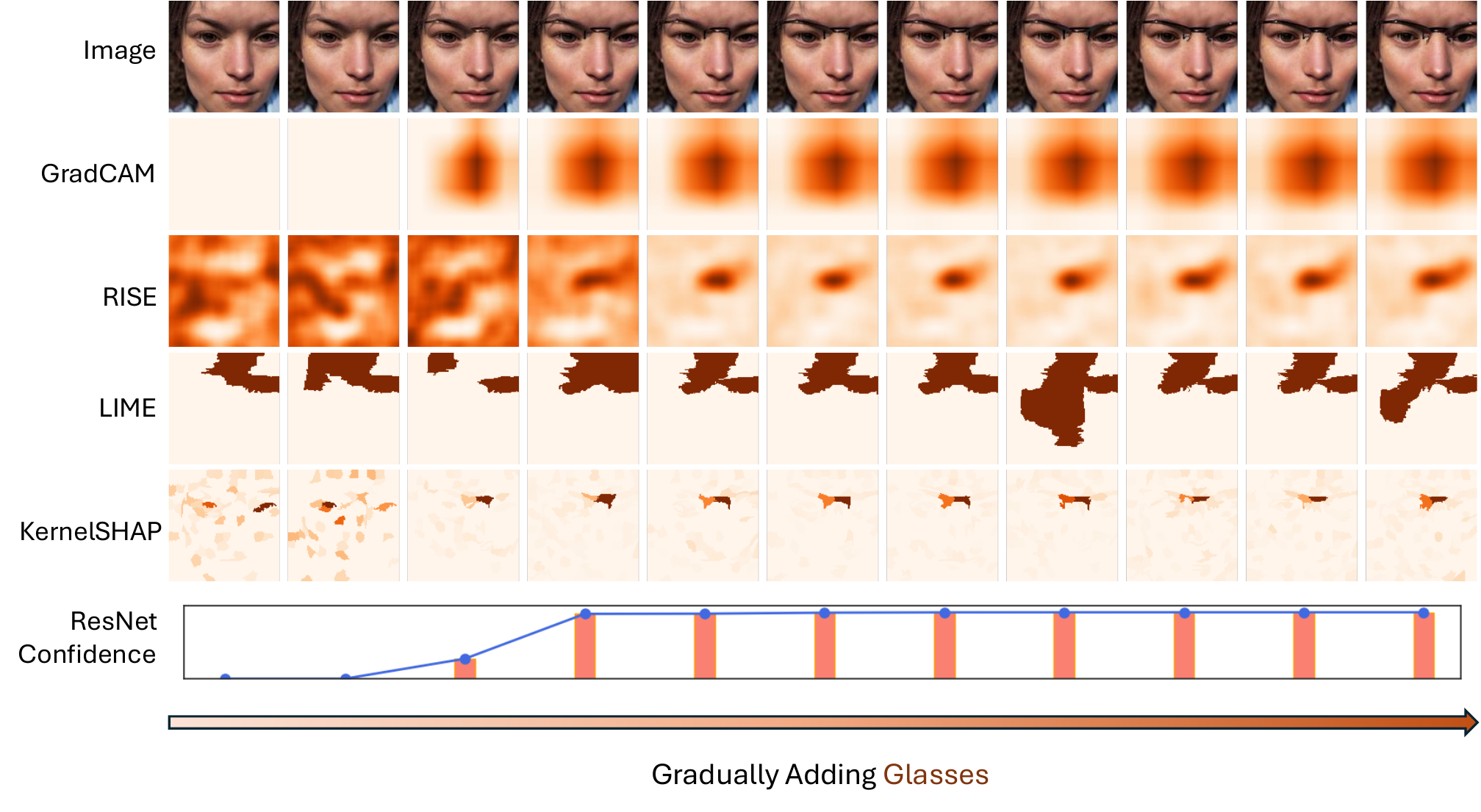}
    \caption{The explanations from explainers and the model confidence regarding semantic variations. Darker areas suggest a higher impact on the model prediction.
    % The saliency maps inferred by the four XAI methods regarding semantic variation. Darker areas suggest a higher impact on the model prediction.
    }
    \label{fig:face-example-24910}
    \end{subfigure}
    \begin{subfigure}[b]{\textwidth}
    \centering
    \includegraphics[width=\textwidth]{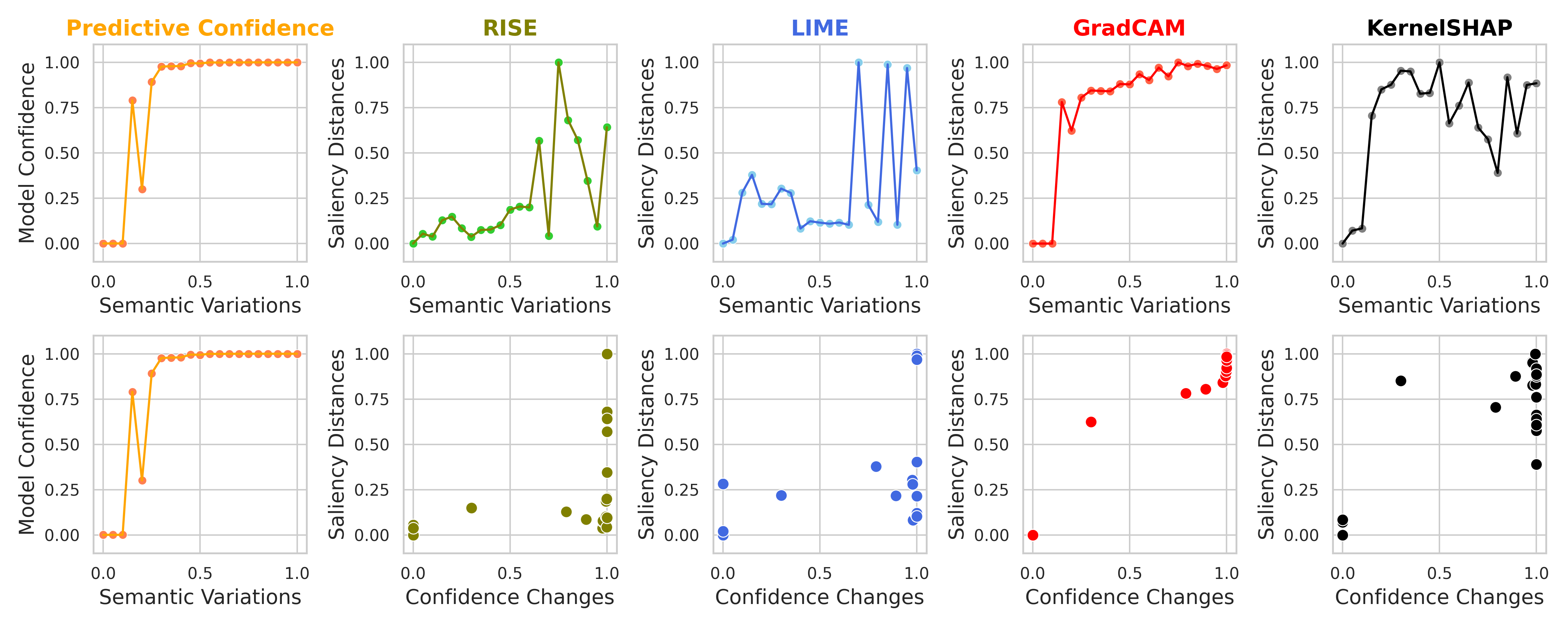}
    \caption{The plots portray the relation among semantic variation, model confidence, and saliency distances. All variables are normalized to \([0,1]\) for better visualization.
    % The relational plots. All variables are normalized to \([0,1]\) for better visualization.
    }
    \label{fig:face-xai-24910}
    \end{subfigure}
    \begin{subfigure}[b]{\textwidth}
    \centering
\begin{tabular*}{\linewidth}{@{\extracolsep{\fill}}cccccc}
Correlation                                   & Metric                           & RISE  & LIME & GradCAM & KernelSHAP \\ \hline
\multicolumn{1}{c|}{\multirow{2}{*}{Kendall}} & \multicolumn{1}{l|}{Wasserstein} & 0.600 & -    & \textcolor{blue}{0.923}   & -          \\
\multicolumn{1}{c|}{}                         & \multicolumn{1}{l|}{MSD}         & 0.779 & -    & \textcolor{blue}{0.923}   & -          \\ \hline
\multicolumn{1}{c|}{\multirow{2}{*}{Pearson}} & \multicolumn{1}{l|}{Wasserstein} & -     & -    & \textcolor{blue}{0.957}   & 0.693      \\
\multicolumn{1}{c|}{}                         & \multicolumn{1}{l|}{MSD}         & -     & -    & \textcolor{blue}{0.964}   & 0.538      \\ \hline
\multicolumn{1}{c|}{\multirow{2}{*}{Spearman}} & \multicolumn{1}{l|}{Wasserstein} & 0.777 & - & \textcolor{blue}{0.983} & - \\
\multicolumn{1}{c|}{}                         & \multicolumn{1}{l|}{MSD}         & 0.923 & -    & \textcolor{blue}{0.983}   & -         
\end{tabular*}
\caption{Statistical correlations between the saliency distances and changes in model confidence. The MSD stands for mean squared deviation, while the Wasserstein is the first Wasserstein distance. The highest correlation coefficient per metric per distance is colored in \textcolor{blue}{blue}. The dash symbol, \textbf{-}, indicates a \(p\geq 0.05\) where it is not evident enough to reject \(H_\mathrm{0}\), i.e., the saliency distances and confidence changes are unlikely correlated.\label{tab:stats-24910}}
    \end{subfigure}
    \caption{\textbf{False positive} example of no-glasses to glasses.
    }
        % An example case where the ResNet \textit{falsely} predicts the emergence of glasses on a girl's face, marking a \textbf{false positive}. To human observers, lenses are missing even in the image furthest to the right. Despite this, all four explainers offer logical rationales: starting from the third column to the left in the saliency maps, the regions of interest are densely around brow ridge and nasal bone, suggesting the model heavily weighs the presence of a glasses frame. The saliency maps contradict the trends of Wasserstein distance metrics. As the model's confidence converges, the distance measurements from all explainers, except those from GradCAM, become stochastic and incoherent, whereas the saliency maps keep their consistency. Furthermore, empirical analysis through relational plots reveals that only GradCAM exhibits certain conformity with the predictor, while the KernalSHAP and RISE behave similarly.
    \label{fig:result-24910}
\end{figure}

\paragraph{False positive}
Figure~\ref{fig:face-example-24910} illustrates a false positive scenario, wherein the model erroneously predicts the presence of glasses where the lenses are missing. All four explainers offer logical rationales: starting from the third column to the left in Figure~\ref{fig:face-example-24910}, the regions of interest are dense around the brow ridge and nasal bone, suggesting the model heavily weighs the presence of a glasses frame. The saliency maps contradict the trends of Wasserstein distance metrics. In Figure~\ref{fig:face-xai-24910}, as the model's confidence converges, the distance measurements from all explainers, except those from GradCAM, become stochastic and incoherent, whereas the saliency maps keep their consistency. Furthermore, empirical analysis through relational plots reveals that only GradCAM exhibits certain conformity with the predictor, while the KernalSHAP and RISE explainers behave similarly. The empirical findings are further substantiated by statistical correlation coefficients in~\ref{tab:stats-24910}, where the discrepancies in GradCAM's explanations demonstrate a stronger correlation with changes in model confidence. In contrast, it is unlikely that LIME is monotonic with the predictor judging by the correlation coefficients.

% \paragraph{False positive}
% \begin{figure}[]
%     \centering
%     \includegraphics[width=\textwidth]{figures/face_dataset/Example/pdfs/24910_stack_full.pdf}
%     \caption{An example case where the ResNet \textit{falsely} predicts the emergence of glasses on a girl's face (a \textbf{false positive}). It is obvious to humans that, lenses are still missing even in the right-most image. The saliency maps of the ResNet \textit{inferred} by the four XAI methods for the presented semantically continuous images are provided accordingly.}
%     \label{fig:face-example-24910}
% \end{figure}

\begin{figure}[]
\vspace{-1em}
    \centering
    \begin{subfigure}[b]{\textwidth}
    \centering
    \includegraphics[width=\textwidth]{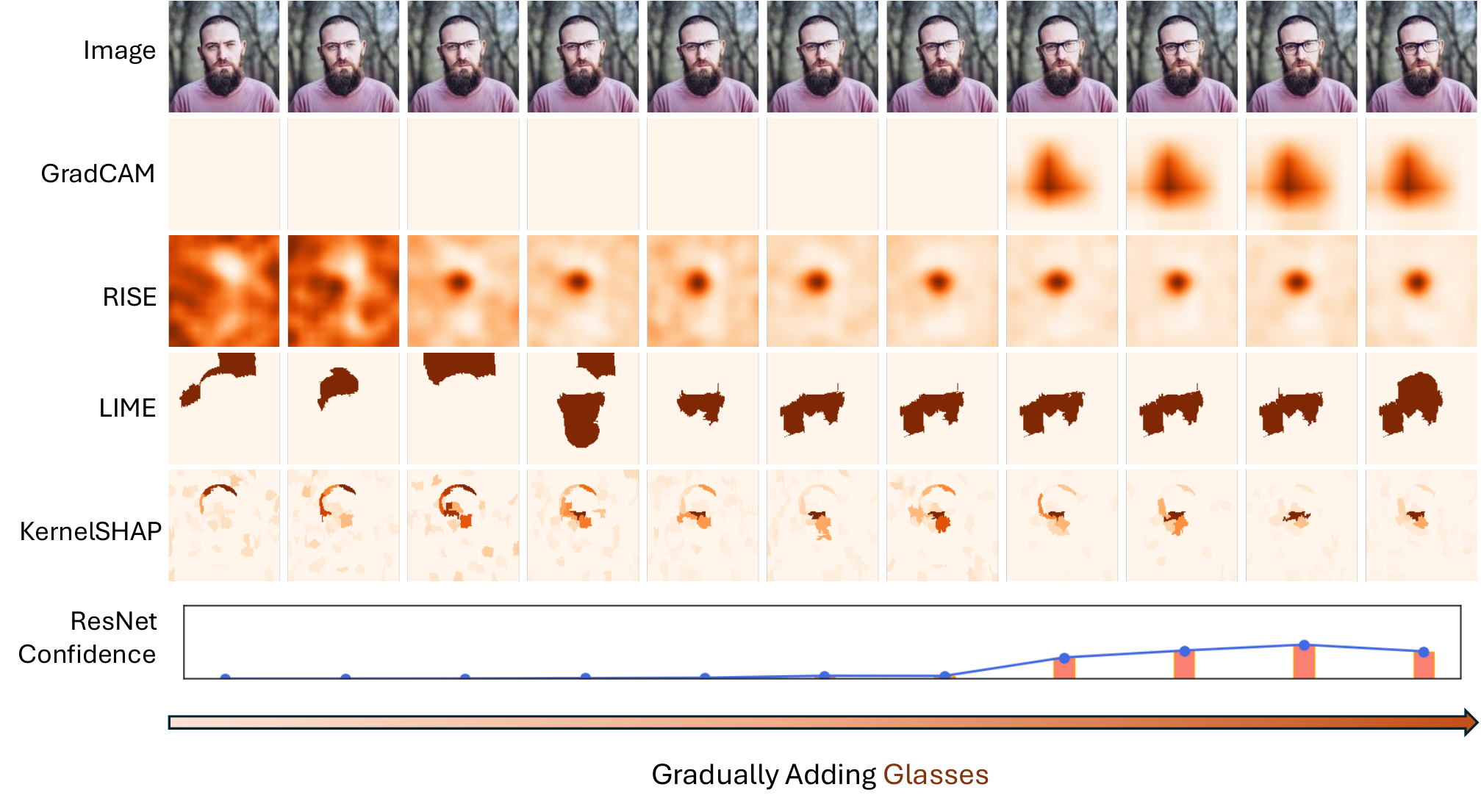}
    \caption{
    The explanations from explainers and the model confidence regarding semantic variations. Darker areas suggest a higher impact on the model prediction.
    % The saliency maps inferred by the four XAI methods regarding semantic variation. Darker areas suggest a higher impact on the model prediction.
    }
    \label{fig:face-example-65553}
    \end{subfigure}
    \begin{subfigure}[b]{\textwidth}
    \centering
    \includegraphics[width=\textwidth]{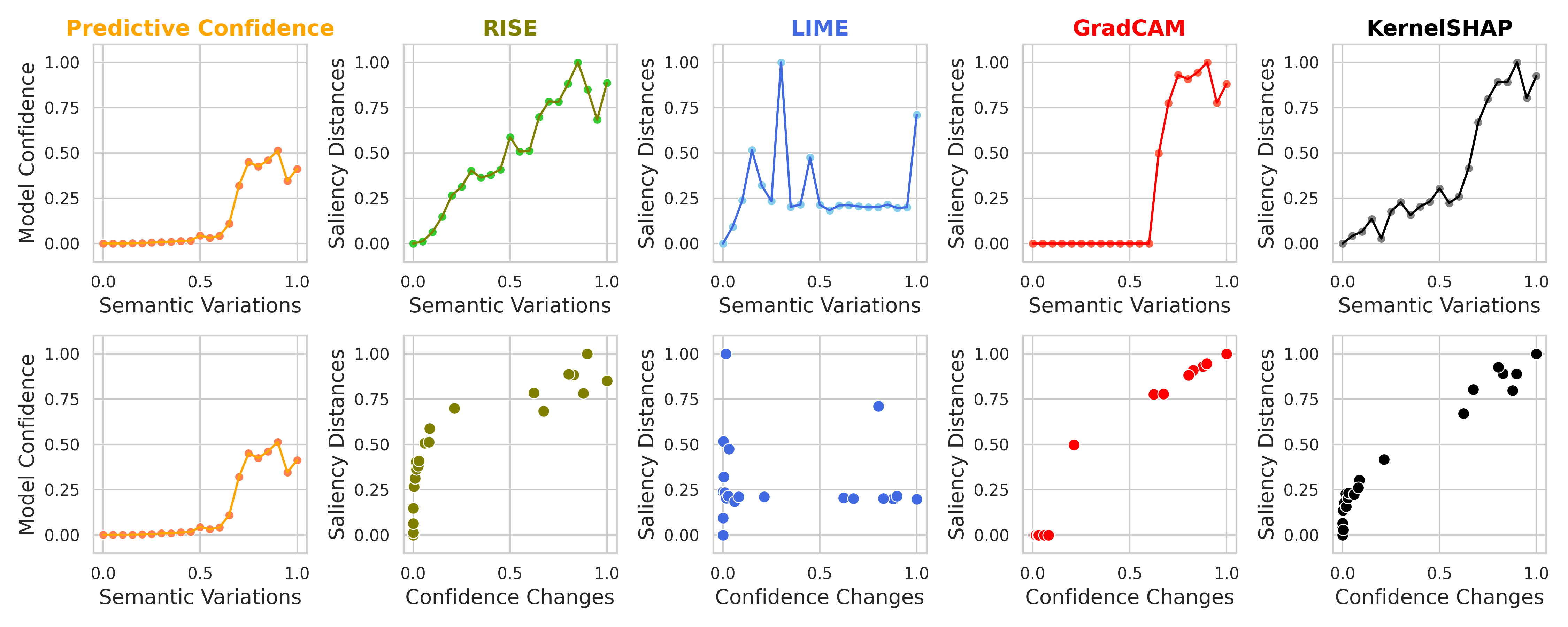}
    \caption{
    The plots portray the relation among semantic variation, model confidence, and saliency distances. All variables are normalized to \([0,1]\) for better visualization. 
    % The relational plots. All variables are normalized to \([0,1]\) for better visualization.
    }
    \label{fig:face-xai-65553}
    \end{subfigure}
    \begin{subfigure}[b]{\textwidth}
    \centering
    \begin{tabular*}{\linewidth}{@{\extracolsep{\fill}}cccccc}
    Correlation           & Metric                           & RISE                         & LIME & GradCAM                      & KernelSHAP                   \\ \hline
    \multicolumn{1}{c|}{} & \multicolumn{1}{l|}{Wasserstein} & 0.884                        & -    & 0.808                        & {\color[HTML]{0000FF} 0.886} \\
    \multicolumn{1}{c|}{\multirow{-2}{*}{Kendall}}  & \multicolumn{1}{l|}{MSD} & 0.863 & 0.421 & 0.795                        & {\color[HTML]{0000FF} 0.884} \\ \hline
    \multicolumn{1}{c|}{} & \multicolumn{1}{l|}{Wasserstein} & 0.866                        & -    & {\color[HTML]{0000FF} 0.984} & 0.981                        \\
    \multicolumn{1}{c|}{\multirow{-2}{*}{Pearson}}  & \multicolumn{1}{l|}{MSD} & 0.927 & 0.52  & {\color[HTML]{0000FF} 0.983} & 0.972                        \\ \hline
    \multicolumn{1}{c|}{} & \multicolumn{1}{l|}{Wasserstein} & {\color[HTML]{0000FF} 0.967} & -    & 0.886                        & 0.958                        \\
    \multicolumn{1}{c|}{\multirow{-2}{*}{Spearman}} & \multicolumn{1}{l|}{MSD} & 0.961 & 0.612 & 0.884                        & {\color[HTML]{0000FF} 0.974}
    \end{tabular*}%
    \caption{Statistical correlations between the saliency distances and changes in model confidence. The MSD stands for mean squared deviation, while the Wasserstein is the first Wasserstein distance. The highest correlation coefficient per metric per distance is colored in \textcolor{blue}{blue}. The dash symbol, \textbf{-}, indicates a \(p\geq 0.05\) where it is not evident enough to reject \(H_\mathrm{0}\), i.e., the saliency distances and confidence changes are likely uncorrelated.\label{tab:stats-65553}}
    \end{subfigure}
    \caption{
    \textbf{False negative} example of no-glasses to glasses.
    % The predictor fails to predict the appearance of complete glasses, resulting in a \textbf{false negative}. 
    % Judging by the saliency maps, RISE and KernelSHAP demonstrate ResNet's growing attention to the right eye. Initially, LIME aligns with KernelSHAP by emphasizing hair, but later, it consistently shows ResNet's focus shifting to both eyes. GradCAM's explanation reveals the predictor's preference for the left cheek as model confidence increases, which plausibly accounts for the model's underperformance. The relational plots further substantiate the tendencies observed in RISE and KernelSHAP. Empirically analysis on relational plots reveals that GradCAM, KernelSHAP, and RISE each display a certain degree of continuity regarding the model confidence, while LIME's inferences show no clear correlation with the predictor's performance.
    }
    \label{fig:result-65553}
\end{figure}

\paragraph{False negative}
In Figure~\ref{fig:result-65553}, we present the analysis of false negatives. The model disbelieves the presence of glasses where they exist. Judging by the saliency maps shown in Figure~\ref{fig:face-example-65553}, RISE and KernelSHAP both exhibit a trend of the ResNet's increasing focus on the right eye. LIME starts to produce a similar inference with KernelSHAP, i.e., highlighting hairs, and it later consistently inferred that the ResNet focuses on both eyes. GradCAM's inference says the predictor is more interested in the left cheek, which reasonably explains the cause of the predictor's underperformance. The tendencies observed in RISE and KernelSHAP are also confirmed through relational plots. Empirically, we find out that GradCAM, KernelSHAP, and RISE each display a certain degree of monotonicity regarding the model confidence, whereas LIME is indifferent. Statistical analysis in Table~\ref{tab:stats-65553} supports our empirical findings. It is hard to precisely rank the continuity of RISE, GradCAM, and KernelSHAP based on these tabular results, however, it is clear that LIME is the explainer with the least explainer continuity.
% \begin{figure}[]
%     \centering
%     \includegraphics[width=\textwidth]{figures/face_dataset/Example/pdfs/65553_stack_full.pdf}
%     \caption{The predictor ResNet fails to predict the appearance of complete glasses (a \textbf{false negative}). The saliency maps of the ResNet \textit{inferred} by the four XAI methods for the presented semantically continuous images are provided accordingly.}
%     \label{fig:face-example-65553}
% \end{figure}

\begin{figure}[]
\vspace{-1em}
    \centering
    \begin{subfigure}[b]{\textwidth}
    \centering
    \includegraphics[width=\textwidth]{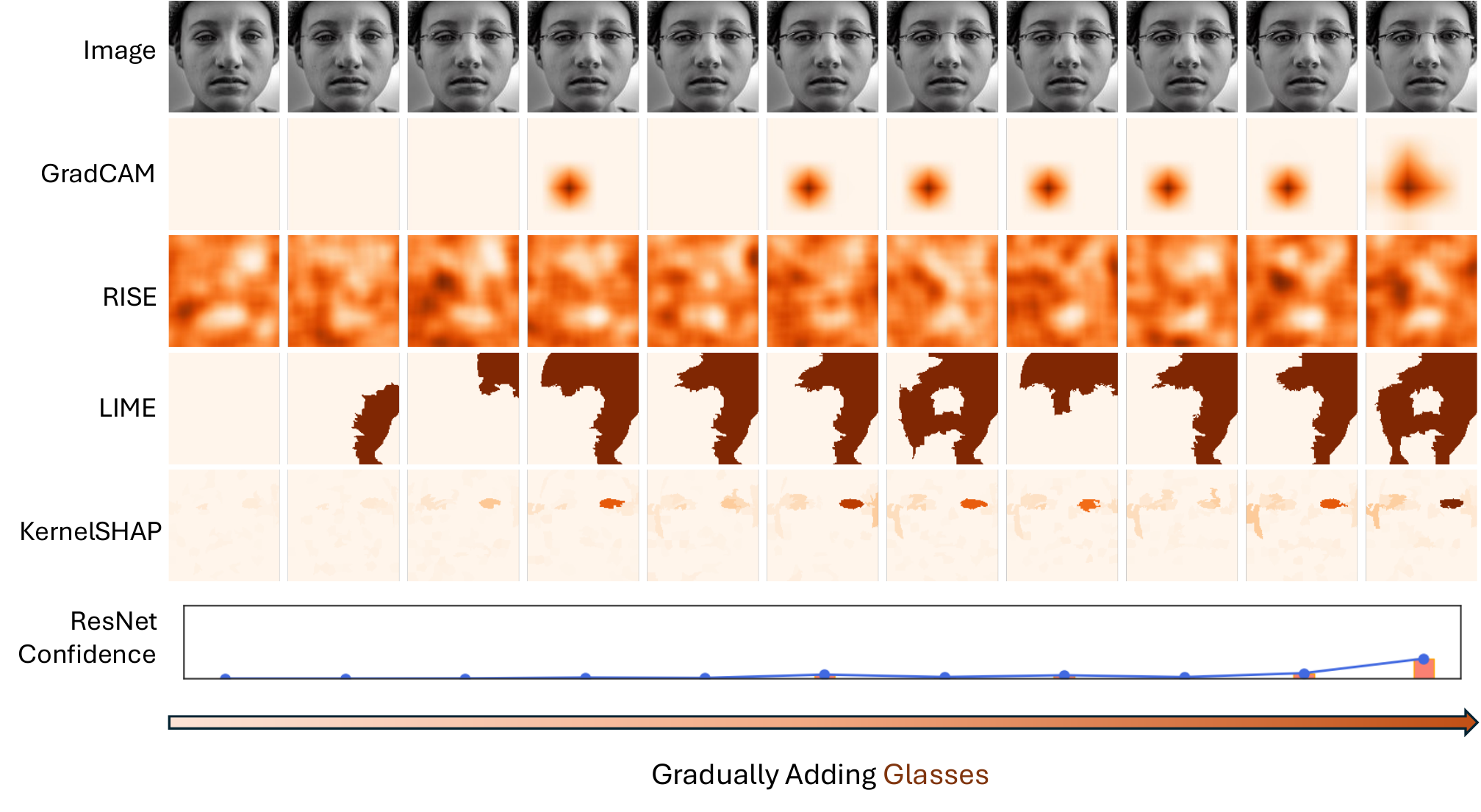}
    \caption{
    The explanations from explainers and the model confidence regarding semantic variations. Darker areas suggest a higher impact on the model prediction.
    % The saliency maps inferred by the four XAI methods regarding semantic variation. Darker areas suggest a higher impact on the model prediction.
    }
    \label{fig:face-example-54991}
    \end{subfigure}
    \begin{subfigure}[b]{\textwidth}
    \centering
    \includegraphics[width=\textwidth]{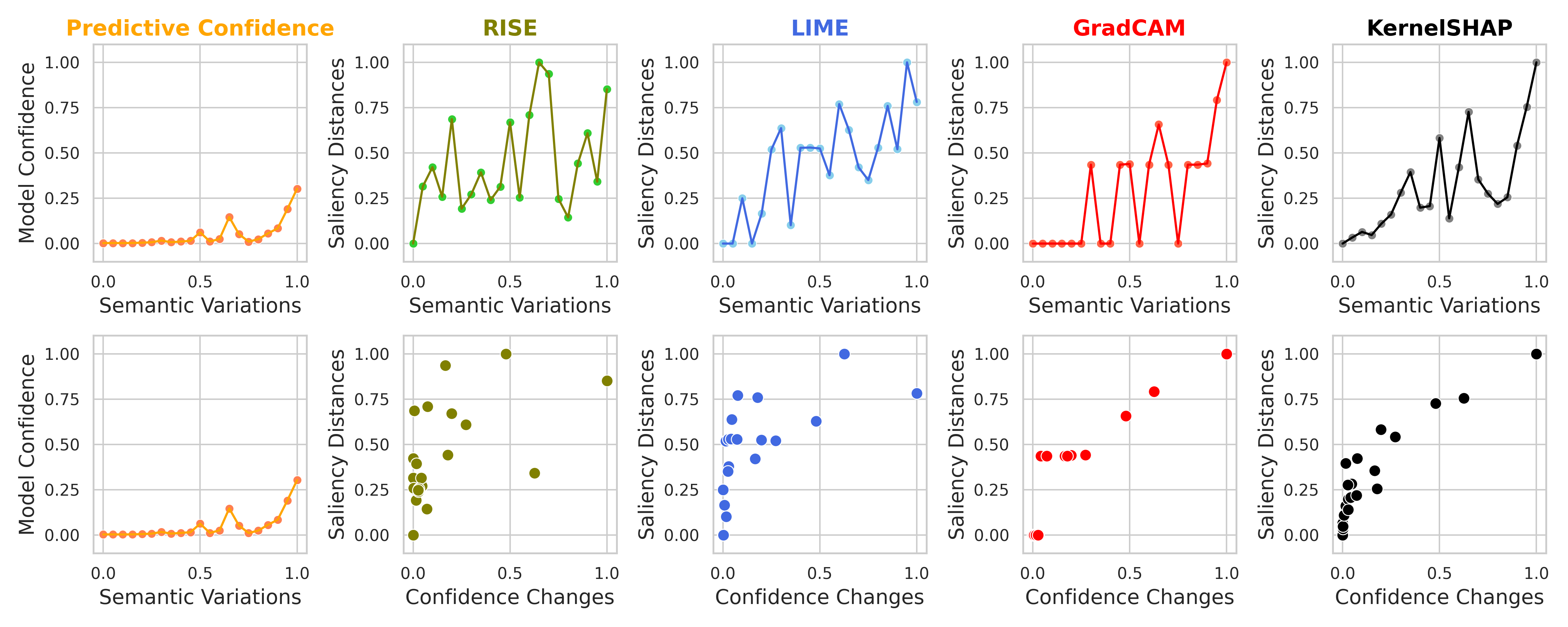}
    \caption{The plots that portray the relation among semantic variation, model confidence, and saliency distances. All variables are normalized to \([0,1]\) for better visualization. 
    % The relational plots. All variables are normalized to \([0,1]\) for better visualization.
    }
    \label{fig:face-xai-54991}
    \end{subfigure}
    \begin{subfigure}[b]{\textwidth}
    \centering
    \begin{tabular*}{\linewidth}{@{\extracolsep{\fill}}cccccc}
    Correlation           & Metric                           & RISE  & LIME  & GradCAM                      & KernelSHAP                   \\ \hline
    \multicolumn{1}{c|}{} & \multicolumn{1}{l|}{Wasserstein} & -     & 0.617 & {\color[HTML]{0000FF} 0.807} & 0.758                        \\
    \multicolumn{1}{c|}{\multirow{-2}{*}{Kendall}}  & \multicolumn{1}{l|}{MSD} & -     & 0.617 & {\color[HTML]{0000FF} 0.847} & 0.695                        \\ \hline
    \multicolumn{1}{c|}{} & \multicolumn{1}{l|}{Wasserstein} & 0.897 & 0.616 & 0.851                        & {\color[HTML]{0000FF} 0.907} \\
    \multicolumn{1}{c|}{\multirow{-2}{*}{Pearson}}  & \multicolumn{1}{l|}{MSD} & 0.616 & 0.616 & 0.832                        & {\color[HTML]{0000FF} 0.957} \\ \hline
    \multicolumn{1}{c|}{} & \multicolumn{1}{l|}{Wasserstein} & 0.48  & 0.801 & {\color[HTML]{0000FF} 0.924} & 0.887                        \\
    \multicolumn{1}{c|}{\multirow{-2}{*}{Spearman}} & \multicolumn{1}{l|}{MSD} & -     & 0.801 & {\color[HTML]{0000FF} 0.939} & 0.842                       
    \end{tabular*}% 
    \caption{Statistical correlations between the saliency distances and changes in model confidence. The MSD stands for mean squared deviation, while the Wasserstein is the first Wasserstein distance. The highest correlation coefficient per metric per distance is colored in \textcolor{blue}{blue}. The dash symbol, \textbf{-}, indicates a \(p\geq 0.05\) where it is not evident enough to reject \(H_\mathrm{0}\), i.e., the saliency distances and confidence changes are likely uncorrelated.\label{tab:stats-54991}}
    \end{subfigure}
    \caption{
    \textbf{True negative} example of no-glasses to glasses.
    % This scenario mirrors that depicted in Figure~\ref{fig:result-54991}, but with a notable difference: ResNet avoids misclassifying images of eyeglasses without lenses (a \textbf{true negative}). Analysis of the heatmaps reveals divergent explanatory behaviours: the heatmaps produced by GradCAM primarily shift between two patterns; RISE's interpretations appear randomly distributed; LIME's explanations concentrate on the upper right segments of the images, albeit with some fluctuation; KernelSHAP consistently emphasizes both eyebrows. Empirical observations from relational plots indicate that GradCAM and KernelSHAP maintain certain explainer continuity regarding predictors, in contrast to the discontinuity exhibited by RISE and LIME despite stable model confidence.
    }
    \label{fig:result-54991}
\end{figure}

\paragraph{True negative}
This scenario in Figure~\ref{fig:result-54991} mirrors that depicted in Figure~\ref{fig:result-54991}, but with a notable difference: ResNet avoids misclassifying images of eyeglasses without lenses (a \textbf{true negative}). Analysis of the heatmaps in Figure~\ref{fig:face-example-54991} reveals divergent explanatory behaviours: the heatmaps produced by GradCAM primarily shift between two patterns; RISE's interpretations appear randomly distributed; LIME's explanations concentrate on the upper right segments of the images, albeit with some fluctuation; KernelSHAP consistently emphasizes both eyebrows. Empirical observations from relational plots in Figure~\ref{fig:face-xai-54991} and statistical results in Table~\ref{tab:stats-54991} indicate that GradCAM and KernelSHAP maintain certain explainer continuity regarding predictors, in contrast to the discontinuity exhibited partially by LIME and particularly by RISE despite low yet stable model confidence.

% \begin{figure}[]
%     \centering
%     \includegraphics[width=\textwidth]{figures/face_dataset/Example/pdfs/54991_stack_full.pdf}
%     \caption{This case is similar to that of Figure~\ref{fig:face-example-24910}. However, this time, the ResNet effectively avoids misclassification when presented with eyeglasses without lenses (a \textbf{true negative}). The saliency maps of the ResNet \textit{inferred} by the four XAI methods for the presented semantically continuous images are provided accordingly.}
%     \label{fig:face-example-54991}
% \end{figure}

% \section{Discussion} @Qi
% \qi{Do we still need this? Or we have already distributed it into results and conclusion.}

Summarizing our analysis across four case studies, visual inspection of saliency maps shows that KernelSHAP delivers the most semantically continuous and informative explanations. RISE and GradCAM follow, ranked second and third, respectively, while LIME is the least informative, with discontinuity between closely adjacent semantics. Regarding metric studies through relational plots and statistical correlations, GradCAM undoubtedly is the most semantically continuous explainer, with KernelSHAP, RISE, and LIME following in descending order.

Besides summarizing findings on an explainer level, we discuss the conformity between qualitative, and our proposed quantitative analysis. Among three of the four studied cases, in Figure~\ref{fig:result-9040}, Figure~\ref{fig:result-24910}, and Figure~\ref{fig:result-65553}, analytical observations on statistical correlation (linearity and monotonicity) between the saliency distances and the changes in model confidence generally match and support our empirical findings on explanations and relational plots. %However, it is necessary to mention 

\FloatBarrier

\section{Conclusions and Outlook}
\label{sec:conclusion}

In this paper, we presented a novel methodology for evaluating semantic continuity for Explainable AI (XAI) methods and subsequently the predictive models. 
Our focus on semantic continuity emphasizes the importance of consistent explanations for similar inputs. We characterize an explainer as semantically continuous if similar inputs, lead to similar model predictions, having similar explanations. 
We explored semantic continuity for image classification tasks, assessing how sequential input changes impact the DL model explanations. 
We explored popular explainers, including LIME, RISE, GradCAM and KernelSHAP. 

We performed an in-depth instance-based analysis for a realistic and complex image binary classification task and different XAI methods. We found that regarding the relational plots and statistical correlations, GradCAM shows to be the most semantically continuous explainer, with KernelSHAP following as a good second. Visual inspection results of saliency maps are mostly in agreement with the proposed qualitative and quantitative semantic continuity measures.

The investigation of semantic continuity extends our understanding of the interpretability of DL models and the capacities of different XAI methods, introducing a crucial dimension to the evaluation of XAI methods.

Numerous promising avenues exist for future research in the intersection of semantic continuity and XAI. First, the proposed metric can be further extended to support different types of deep learning tasks beyond image classification and also extend to other domains, such as text or speech. Additionally, exploring the impact of semantic continuity on user trust and acceptance of DL models and XAI methods can provide insights into the practical implications of our findings. To quantify the semantic continuity, different metrics could be explored such as Distance correlation \cite{szekely2007measuring}.
In conclusion, we hope that our comprehension of semantic continuity and its nuanced implications not only enhances the evolution of Explainable AI methods but also supports the usage of deep learning across diverse domains.

\subsubsection{Disclosure of Interests.} The authors have no competing interests to declare that are relevant to the content of this article.

% \section*{Acknowledgments}
% 
\section*{Acknowledgments}
We would like to thank the Lorentz Center for organizing the Workshop; ICT with Industry, which led to the collaboration and this work.
This publication is part of the project XAIPre (with project number 19455) of the research program Smart Industry 2020 which is (partly) financed by the Dutch Research Council (NWO). Funding for Osman Mutlu in this research has been provided by the European Union’s Horizon Europe research and innovation programme EFRA [grant number 101093026].

\bibliographystyle{splncs04}
\bibliography{main.bib}

\appendix

\end{document}